\documentclass[10pt,twocolumn,letterpaper]{article}
\usepackage[pagenumbers]{cvpr}
\usepackage{subcaption}
\usepackage{tabularx, booktabs}
\usepackage{longtable}

\newcommand{\method}[1]{\texttt{RadDiff}}
\newcommand{\benchmark}[1]{\texttt{RadDiffBench}}
\newcommand{\Task}[1]{Set difference captioning}
\newcommand{\task}[1]{set difference captioning}

\newcommand{\seta}{$\mathcal{R}_A$}
\newcommand{\setb}{$\mathcal{R}_B$}

\definecolor{cvprblue}{rgb}{0.21,0.49,0.74}
\usepackage[pagebackref,breaklinks,colorlinks,allcolors=cvprblue]{hyperref}

\def\paperID{N/A}
\def\confName{CVPR}
\def\confYear{2026}

\title{RadDiff: Describing Differences in Radiology Image Sets with Natural Language}

\author{\textbf{Xiaoxian Shen\thanks{Equal contribution.~$^\dagger$Correspondence to: \tt{\{yuhuiz,syyeung\}@stanford.edu}.~\textnormal{Code is available} \href{https://github.com/yuhui-zh15/RadDiff}{here}.} \quad Yuhui Zhang$^{*\dagger}$ \quad Sahithi Ankireddy \quad Xiaohan Wang \quad Maya Varma} \\ \textbf{Henry Guo \quad Curtis Langlotz \quad Serena Yeung-Levy$^\dagger$} \\
Stanford University\\
}

\begin{document}

\maketitle

\begin{abstract}

Understanding how two radiology image sets differ is critical for generating clinical insights and for interpreting medical AI systems. We introduce \method{}, a multimodal agentic system that performs radiologist-style comparative reasoning to describe clinically meaningful differences between paired radiology studies. \method{} builds on a proposer-ranker framework from VisDiff, and incorporates four innovations inspired by real diagnostic workflows: (1) medical knowledge injection through domain-adapted vision-language models; (2) multimodal reasoning that integrates images with their clinical reports; (3) iterative hypothesis refinement across multiple reasoning rounds; and (4) targeted visual search that localizes and zooms in on salient regions to capture subtle findings. To evaluate \method{}, we construct \benchmark{}, a challenging benchmark comprising 57 expert-validated radiology study pairs with ground-truth difference descriptions. On \benchmark{}, \method{} achieves 47\% accuracy, and 50\% accuracy when guided by ground-truth reports, significantly outperforming the general-domain VisDiff baseline. We further demonstrate \method{}’s versatility across diverse clinical tasks, including COVID-19 phenotype comparison, racial subgroup analysis, and discovery of survival-related imaging features. Together, \method{} and \benchmark{} provide the first method-and-benchmark foundation for systematically uncovering meaningful differences in radiological data.

\end{abstract}    
\section{Introduction}
\label{sec:intro}

\begin{figure}
    \centering
    \includegraphics[width=1\linewidth]{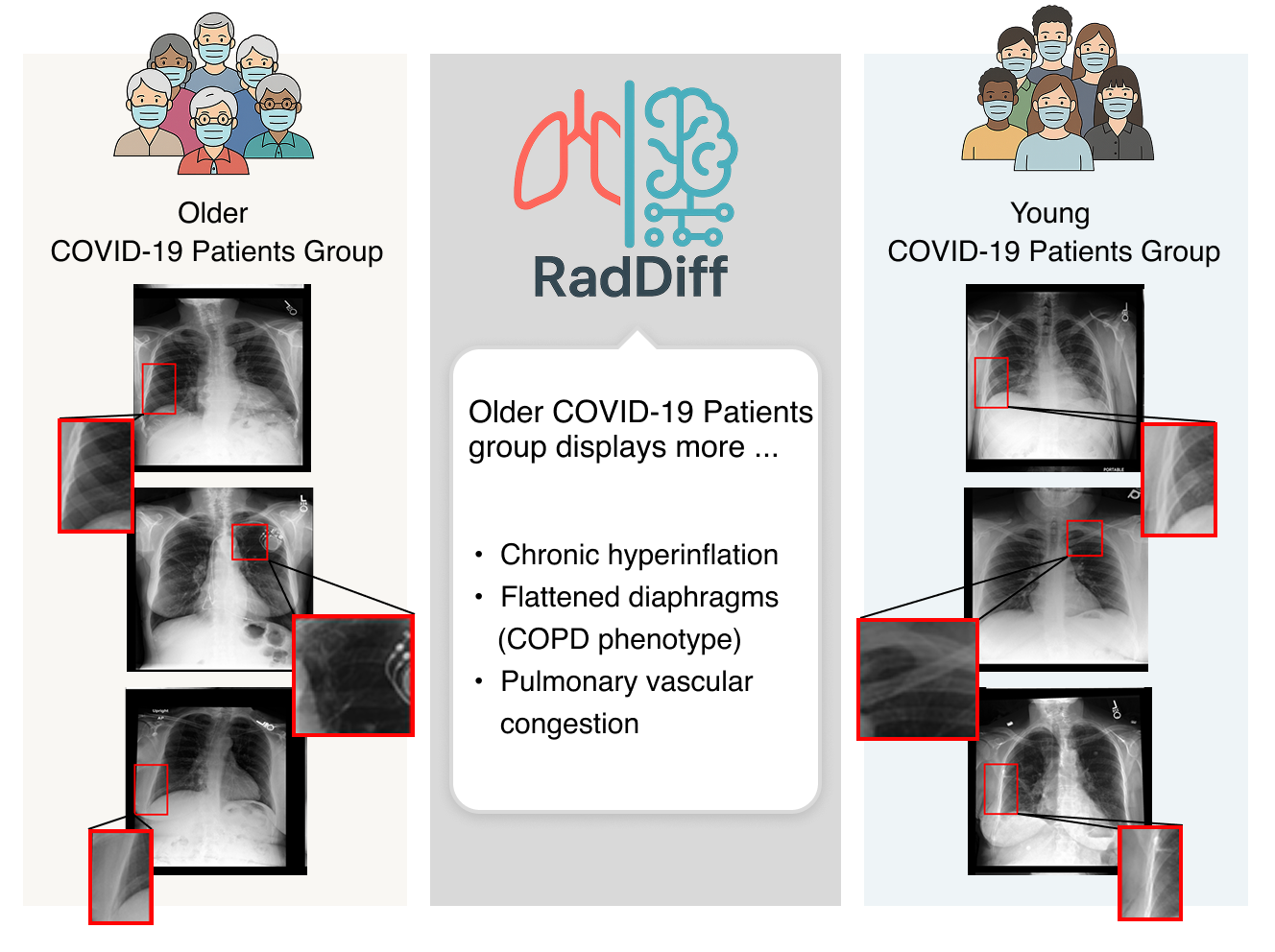}
    \vspace{-2em}
    \caption{\method{} is designed to identify the differences between two groups of radiology images. In this example, older COVID-19 patients display more findings than younger COVID-19 patients.}
    \vspace{-1.5em}
    \label{fig:intro}
\end{figure}

\begin{figure*}[!tb]
    \centering
    \includegraphics[width=1\linewidth]{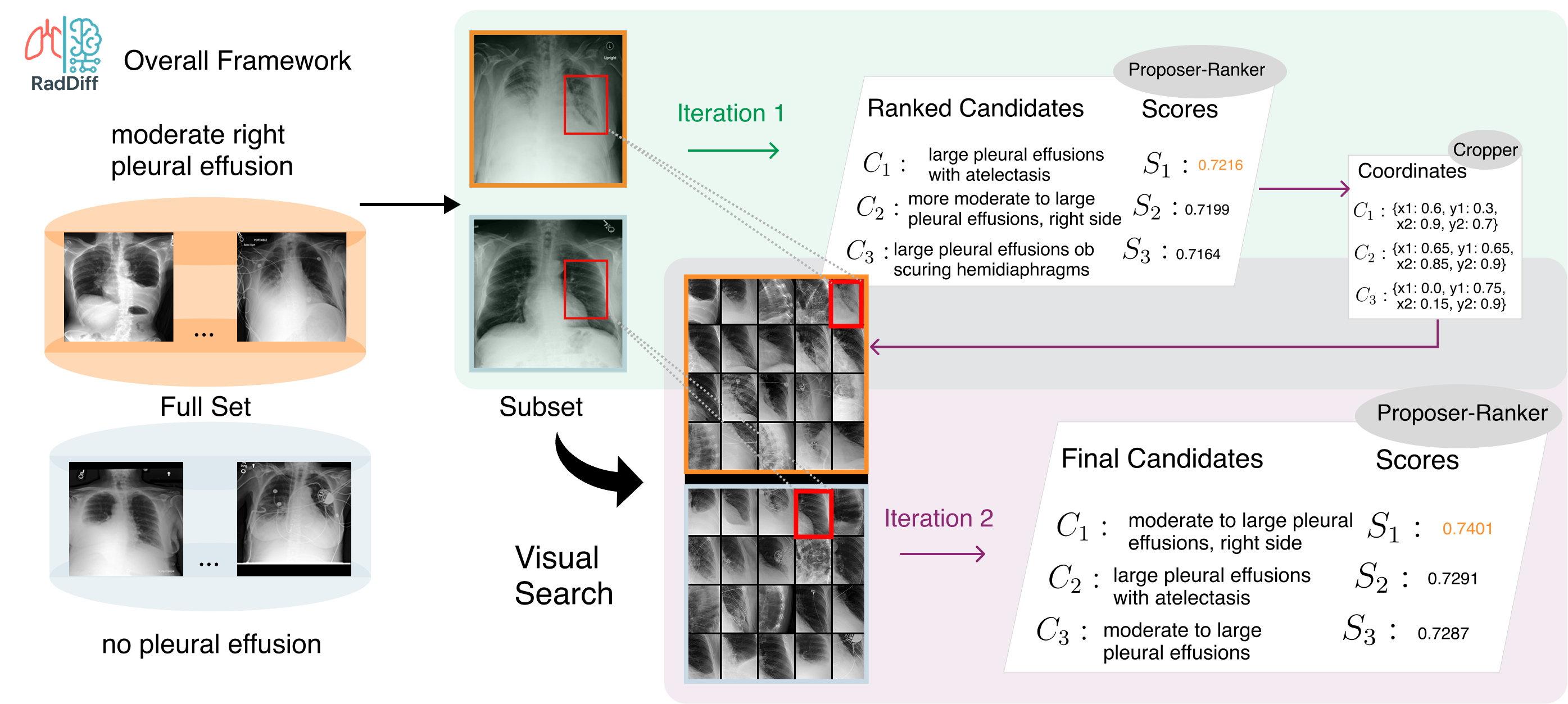}
    \vspace{-2em}
    \caption{\textbf{\method{} algorithm.} To solve the challenging task of identifying differences between two large sets consisting of thousands of images, \method{} leverages the proposer–ranker framework from VisDiff, which first generates candidate differences from subsets and then ranks them based on a saliency score reflecting differences between the full sets. \method{} incorporates four improvements to enhance performance: (1) medical knowledge injection through domain-adapted vision-language models; (2) multimodal reasoning that integrates images with their clinical reports; (3) iterative hypothesis refinement across multiple reasoning rounds; and (4) targeted visual search that localizes and zooms in on salient regions to capture subtle findings.}
    \vspace{-1em}
    \label{fig:main_raddiff}
\end{figure*}

What are the distinct phenotypes of young versus elderly COVID-19 patients~\cite{okoye2022computed}? What characteristics separate patients who survive pneumonia from those who do not~\cite{kim2023deep}? Why can image classifiers accurately identify patient race from radiology images~\cite{gichoya2022ai, lotter2024acquisition}? Understanding such questions is essential for generating new clinical insights and for debugging medical AI models. However, answering them remains challenging even for experts, as it requires careful, time-consuming inspection and reasoning over two large cohorts of radiology images.

In this work, we introduce \method{}, a multimodal agent that automatically generates clinically meaningful differences between two large cohorts of radiology images. \method{} builds on the VisDiff~\cite{dunlap2024describing} proposer–ranker framework, which first uses vision–language models (VLMs)~\cite{llava1.5} to generate image captions, and large language models (LLMs)~\cite{gpt4} to propose candidate differences based on image captions, and then ranks them using multimodal embeddings~\cite{radford2021learning} based on a saliency score reflecting how strongly the cohorts differ. However, directly applying VisDiff to medical imaging yields poor performance. Radiographs contain subtle and fine-grained findings that require domain expertise, anatomical priors, and joint reasoning across multiple structures—capabilities that VisDiff’s text-only reasoning cannot provide. Moreover, VisDiff performs single-pass reasoning, whereas radiologists iteratively search, compare, and refine hypotheses across multiple rounds.

To address these limitations, we emulate radiologist-style comparative reasoning and introduce four methodological advances in \method{}. (1) Medical knowledge injection: We adapt domain-specific VLMs—including the CheXagent~\cite{chen2024chexagent} vision–language model and the CheXzero~\cite{tiu2022expert} CLIP-style model—and incorporate medical instructions to ensure recognition of clinically relevant entities. (2) Multimodal reasoning: \method{} processes both radiology images and paired clinical notes, enabling joint interpretation of spatial visual cues and caption-level contextual information. (3) Iterative refinement: Candidate differences proposed in earlier rounds serve as contextual evidence for subsequent rounds, mirroring how radiologists revisit and update hypotheses. (4) Targeted visual search: A visual-search module localizes salient regions for each candidate difference and extracts fine-grained image patches, allowing \method{} to attend to subtle patterns that would otherwise be overlooked.

We evaluate \method{} on \benchmark{}, a new radiologist-verified benchmark we construct to support method development. \benchmark{} contains 57 expert-validated paired cohorts derived from MIMIC-CXR \cite{li2023mimicit}, each with ground-truth descriptions of clinically relevant differences. Benchmark construction proceeds in two stages. (1) We use LLMs to propose 150 clinically meaningful cohort pairs (e.g., patients with vs. without pleural effusion). Radiologists validate these proposals, resulting in 57 final groups, and assign easy, medium, and hard difficulty levels. (2) We then collect images for each cohort using clinical reports as a proxy label. Because no reliable open-vocabulary classifier exists for radiographs, we classify images in the text domain: we first perform BM25-based retrieval using report text, then use an LLM to confirm that each retrieved report aligns with the target description. This yields approximately 600 images per cohort. Radiologists perform a final review to ensure benchmark quality.

On \benchmark{}, \method{} significantly outperforms the general-domain VisDiff baseline, improving accuracy from 2\% to 47\%---a 45-point gain. Ablations confirm that each component—medical knowledge injection, multimodal reasoning, iterative refinement, and targeted visual search—substantially contributes to the improvement, especially on the hardest subsets requiring fine-grained spatial understanding and complex reasoning. When provided with ground-truth reports, \method{} reaches 50\% accuracy, suggesting additional gains are possible when high-quality clinical text is available.

We further apply \method{} to real-world clinical discovery and model analysis tasks. \method{} identifies coherent cohort-level distinctions across the motivating scenarios. For example, it reveals that younger COVID-19 patients tend to show more acute infection whereas older patients display chronic structural changes; that low-mortality cohorts exhibit fewer medical devices and milder parenchymal disease than high-mortality cohorts; and that models differentiate racial groups not by anatomy, but via acquisition-related confounders, yielding underdiagnosis bias, notably, detecting more abnormalities for White patients relative to others. These findings align with known clinical patterns \cite{okoye2022computed, kim2023deep, gichoya2022ai, lotter2024acquisition} while also highlighting additional insights potentially overlooked in manual review.

In summary, \method{} provides a practical and general tool for generating clinically informative differences between large radiology cohorts. We introduce key methodological improvements, validate their effectiveness on \benchmark{}, and demonstrate \method{}’s utility in real-world clinical investigations. Together, our work offers the first principled framework for describing population-level differences in radiology images and opens new avenues for scientific discovery, fairness analysis, and interpretable cohort-level comparison in medical imaging.

\vspace{-0.3em}

\section{Related Work}
\label{sec:formatting}

\noindent\textbf{Vision-language models.} Vision-language models (VLMs) represent a broad class of models that integrate visual and textual inputs to enable rich multimodal representation and generation. Broadly, they can be categorized into \emph{embedding-based} and \emph{generative} models. Embedding-based contrastive models such as CLIP learn aligned representation spaces for vision and language~\cite{radford2021learning,girdhar2023imagebind,zhai2023sigmoid}, whereas generative multimodal language models (MLLMs) such as GPT are capable of reasoning over complex visual inputs~\cite{gpt4,llava1.5,li2023blip}. Recently, VLMs have been further composed into agentic systems to address more complex tasks~\cite{wang2024videoagent,shen2023hugginggpt,gupta2023visual,suris2023vipergpt}---for example, VisDiff~\cite{dunlap2024describing} combines CLIP and MLLMs to detect dataset-level differences. In this work, we adapt the VisDiff algorithm to the radiology domain and introduce four key enhancements to make it effective in this setting.

\noindent\textbf{Radiology applications of VLMs.} Recent efforts have extended VLMs to medical imaging, with a primary focus on chest X-ray interpretation. Prominent embedding-based VLMs include CheXzero~\citep{tiu2022expert}, BioVIL~\citep{Boecking_2022}, GLoRIA~\citep{huang2021gloria}, ConVIRT~\citep{zhang2022contrastivelearningmedicalvisual}, and MGCA~\citep{wang2022multigranularity}. More recent work has introduced generative VLMs that couple strong medical image encoders with large language models, such as CheXagent~\citep{chen2024chexagent}, MAIRA-2~\citep{bannur2024maira2groundedradiologyreport}, and MedGemma~\citep{sellergren2025medgemmatechnicalreport}. These systems demonstrate strong performance across clinically relevant tasks, including visual question answering, disease classification, longitudinal analysis, medical reasoning, and radiology report generation. In this work, we propose a new radiology task: describing differences between sets of radiology images using natural language---a practically important yet technically challenging problem. To support this task, we introduce \benchmark{} and \method{}, a benchmark and a method that provides a foundation for solving this task.

\section{Problem Formulation}

In this section, we first formulate our task, then describe the construction of \benchmark{}, a challenging benchmark consisting of 57 pairs of expert-validated differences, and finally explain the evaluation framework.

\begin{figure}[!tb]
    \centering
    \includegraphics[width=1\linewidth]{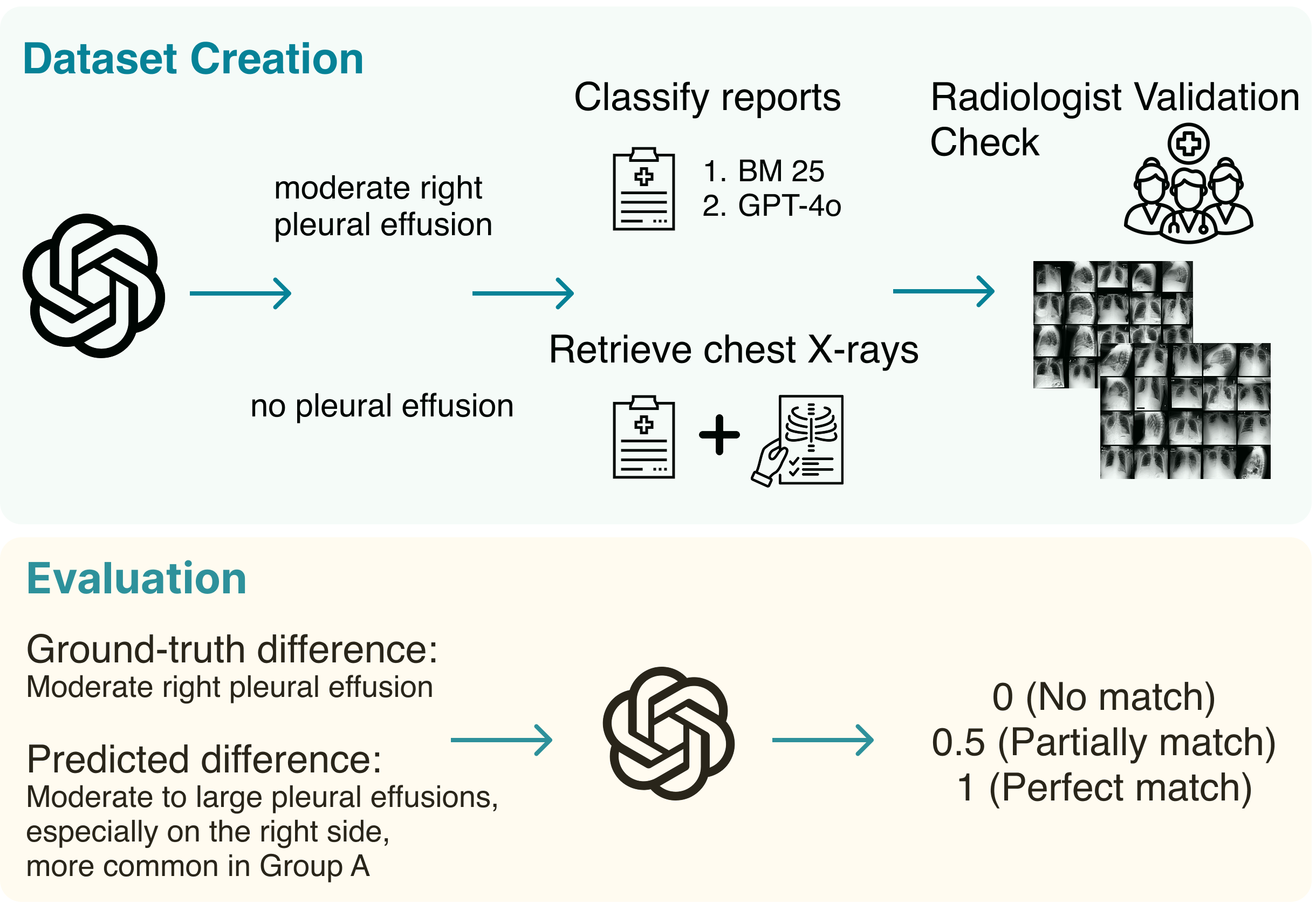}
    \vspace{-2em}
    \caption{\textbf{\benchmark{} creation and evaluation.} \benchmark{} is created in two stages: we first use LLMs to propose clinically meaningful cohort pairs, and then classify images into each pair by using an LLM to categorize clinical reports as a proxy for the image labels. For evaluation, we use an LLM to assign a three-level score representing the similarity between the predicted difference and the ground-truth difference. }
    \label{fig:placeholder}
    \vspace{-1.5em}
\end{figure}

\subsection{Task Description}

Given two sets of radiology images, \seta{} and \setb{}, our goal is to identify the anatomical and pathological differences that distinguish one group from the other. In particular, we aim to describe features that are more prevalent in \seta{} than in \setb{}. 

Formally, the task is defined as learning a mapping
\begin{equation}
    G: (\mathcal{R}_A, \mathcal{R}_B) \rightarrow \mathcal{C}_{A>B},
\end{equation}
where $\mathcal{C}_{A>B} = \{ c_1, c_2, \ldots, c_n \}$ denotes a set of textual difference descriptions capturing findings more common in group $A$ than in group $B$.

In practice, \seta{} and \setb{} may be very large, containing thousands of images, and the description space $\mathcal{C}_{A>B}$ is open-ended, underscoring the difficulty of the task.

\subsection{Benchmark}

Since this is a novel task and no existing benchmark is available, we construct \benchmark{} to enable systematic evaluation and development of our system. The construction of \benchmark{} follows a two-stage pipeline.

In the first stage, we use GPT-4o~\citep{gpt4} to propose hypothetical differences between paired image sets (e.g., ``moderate right pleural effusion'' vs. ``no pleural effusion''). Because this task requires domain-specific expertise, we provide GPT-4o with sampled radiology reports from the MIMIC-CXR dataset~\citep{Johnson2019} to improve the medical relevance of the generated differences. GPT-4o produces 150 candidate difference groups, and radiologists validate their clinical usefulness and assign difficulty levels (easy, medium, hard), resulting in 57 finalized differences.

In the second stage, we classify each chest X-ray in the MIMIC-CXR dataset~\citep{Johnson2019} into set A, set B, or neither. This is challenging because no reliable open-vocabulary chest X-ray classifier currently exists. Fortunately, MIMIC-CXR provides ground-truth radiology reports associated with each image, allowing us to use the report text as a proxy for image-based classification. Determining whether a finding is present in text is far easier than in images. We first use the BM25 algorithm~\cite{robertson2009probabilistic} to retrieve reports with similar keywords, then apply GPT-4o-mini~\cite{gpt4} for fine-grained semantic matching. Radiologist validation shows that this process achieves near-perfect accuracy.

In summary, \benchmark{} contains 57 expert-validated differences spanning multiple difficulty levels. Table~\ref{tab:dataset_stats} presents the final benchmark statistics.

\begin{table}[!tb]
    \small
    \centering
\begin{tabular}{lcccc}
\toprule
\textbf{} & \textbf{Total} & \textbf{Easy} & \textbf{Medium} & \textbf{Hard} \\
\midrule
\textbf{\#Pairs} & 57 & 23 & 21 & 13 \\
\textbf{Mean \#CXRs Per Pair} & 614 & 614 & 607 & 625 \\
\bottomrule
\end{tabular}
\vspace{-1em}
    \caption{\textbf{Statistics of \benchmark{}.}}
    \label{tab:dataset_stats}
    \vspace{-1.5em}
\end{table}

\subsection{Evaluation}

For evaluation, algorithms generate a list of descriptions $\mathcal{C}_{A>B}$ for each pair (\seta{}, \setb{}), which we compare to the ground-truth description $c^*$ provided in \benchmark{}. To measure the similarity between each $c_i$ and $c^*$, we use GPT-4.1-nano, prompted to categorize each proposed difference as a match (1), partial match (0.5), or no match (0). Prior work demonstrates strong alignment between LLM-based and human evaluations~\citep{dunlap2024describing, dubois2023alpacafarm}. 

We report Acc@1/5/N, which measures whether the ground-truth description appears within the top 1, 5, or N ranked generated descriptions.

\section{Method}

Our task—identifying differences between large radiology image sets—requires reasoning over thousands of images, which is challenging even for human experts. To address this, we adapt a proposer–ranker framework and introduce four enhancements inspired by radiologist workflow: (1) Knowledge Injection, (2) Multimodal Reasoning, (3) Iterative Refinement, and (4) Visual Search.

\subsection{Proposer + Ranker Framework}

Because no existing model can reliably reason over two large sets containing thousands of images, we adopt the proposer–ranker framework introduced by VisDiff~\cite{dunlap2024describing}. The proposer generates candidate differences, and the ranker assigns each candidate a score measuring how salient that difference is between the two sets.

\textbf{Proposer.} The proposer samples random subsets $\mathcal{X}_A \subset \mathcal{R}_A$ and $\mathcal{X}_B \subset \mathcal{R}_B$, and generates candidate differences based on these subsets. In practice, we set $|\mathcal{X}_A| = |\mathcal{X}_B| = 20$. In VisDiff, the proposer incorporates an MLLM-based image captioner~\cite{llava1.5} that first generates image captions, after which an LLM\footnote{We use GPT-4.1-nano in our experiments.}~\cite{gpt4} proposes candidate differences using those captions. This design reflects the substantial reasoning capabilities required at this stage.

\textbf{Ranker.} Since the proposer observes only a small subset of images, its candidates may not reflect the most representative differences. The ranker evaluates each candidate difference $c \in \mathcal{C}_{A>B}$ against the full datasets \seta{} and \setb{}. It computes a discriminative score  
\[
s_c = \mathbb{E}_{x \in \mathcal{R}_A} v(x, c) - \mathbb{E}_{x \in \mathcal{R}_B} v(x, c),
\]
where $v(x, c)$ measures how well an image $x$ aligns with candidate difference $c$. We use a CLIP model~\cite{radford2021learning} due to its strong cross-modal concept alignment, defining $v(x, c)$ as the cosine similarity between image embeddings $e_x$ and text embeddings $e_y$.

\subsection{Methodology Improvements}

\begin{table*}[!tb]
\small
\centering
\setlength\tabcolsep{4pt}
\begin{tabular}{lccccccccc}
\toprule
\textbf{Method} &
\multicolumn{3}{c}{\textbf{Average}} &
\multicolumn{2}{c}{\textbf{Easy}} &
\multicolumn{2}{c}{\textbf{Medium}} &
\multicolumn{2}{c}{\textbf{Hard}} \\
 & Acc@1 & Acc@5 & Acc@N & Acc@1 & Acc@5 & Acc@1 & Acc@5 & Acc@1 & Acc@5 \\
\midrule
VisDiff & 0.0175 & 0.0351 & 0.2895 & 0.0435 & 0.0435 & 0.0000 & 0.0238 & 0.0000 & 0.0385 \\
+ Knowledge Injection (CheXagent) & 0.0965 & 0.3070 & 0.7807 & 0.1739 & 0.5000 & 0.0238 & 0.2143  & 0.0769 & 0.1154 \\
+ Knowledge Injection (CheXzero) & 0.2895 & 0.5789 & 0.7807 & 0.4130 & 0.6522 & 0.2857 & 0.7381 & 0.0769 & 0.1923 \\
+ Knowledge Injection (Domain Prompt) & 0.2982 & 0.6228 & 0.8421 & 0.3261 & 0.6522 & 0.4048 & \textbf{0.7857} & 0.0769 & 0.3077 \\
+ Multimodal Reasoning (Joint Image \& Text) & 0.3333 & 0.6316 & 0.8684 & 0.4565 & 0.7174 & 0.4048 & \textbf{0.7857} & 0.0000 & 0.2308 \\
+ Iterative Refinement (Top 5) & 0.4386 & \textbf{0.6930} & \underline{0.8947} & \underline{0.5870} & 0.7391 & 0.4524 & 0.7381 & 0.1538 & \textbf{0.5385} \\
+ Iterative Refinement (Top 10) & \underline{0.4561} & 0.6579 & 0.8596 & 0.5000 & \underline{0.7609} & \textbf{0.5714} & \underline{0.7619} & \underline{0.1923} & 0.3077 \\
\method{} (+ Visual Search) & \textbf{0.4737} & \underline{0.6754} & \textbf{0.9035} & \textbf{0.6087} & \textbf{0.7826} & \underline{0.4762} & \textbf{0.7857} & \textbf{0.2308} & \underline{0.3077} \\
\midrule
Groundtruth Reports & 0.3772 & 0.7807 & 0.9737 & 0.3913 & 0.6957 & 0.4762 & 0.9524 & 0.1923 & 0.6538 \\
+ Iterative Refinement (Top 10) & 0.5088 & 0.7895 & 0.9123 & 0.5217 & 0.7609 & 0.5476 & 0.9048 & 0.4231 & 0.6538 \\
\bottomrule
\end{tabular}
\vspace{-1em}
\caption{\textbf{\method{} results on \benchmark{}.}  
\method{} achieves strong performance on \benchmark{}, attaining 47.37\% top-1 accuracy—a substantial improvement over the general-domain VisDiff baseline. These gains result from the combined contributions of knowledge injection, multimodal reasoning, iterative refinement, and visual search. \textbf{Bolded} values indicate the best results, and \underline{underlined} values indicate the second best.}
\label{tab:main_result}
\vspace{-1.5em}
\end{table*}

We extend the proposer–ranker framework with four improvements motivated by radiologist diagnostic reasoning:

\textbf{Knowledge Injection.} VisDiff uses general-domain MLLM and CLIP as proposer and ranker, which lack the specialized radiology knowledge needed for this task. We incorporate domain-specific models—CheXagent~\citep{chen2024chexagent} for caption generation and CheXzero~\citep{tiu2022expert} for ranking. These models are fine-tuned on medical data and encode detailed chest X-ray knowledge. We additionally refine prompts for the radiology domain. This injects essential prior knowledge into the system.

\textbf{Multimodal Reasoning.} VisDiff performs reasoning solely on text, providing only image captions to the proposer. In radiology, however, fine-grained visual cues are difficult to fully capture in language yet critical for decision making. We therefore enable multimodal reasoning by providing both generated captions and images organized into grids to the proposer, yielding more faithful and clinically meaningful difference proposals.

\textbf{Iterative Refinement.} Radiologists rarely reach conclusions in a single pass; they form, test, and revise hypotheses through iterative comparison and reasoning. To emulate this, we introduce an iterative refinement process. After the initial proposer–ranker cycle, the top $k$ differences with the highest scores are fed back as contextual input for the next proposal round. Each iteration conditions the model on previously identified differences, improving coherence and depth of analysis. We explore different values of $k$ (e.g., $k=5$ or $10$) and iteration rounds $r$ (e.g., $r=2$ or $3$) to balance refinement with diversity; too many iterations can cause redundancy, while too few may limit reasoning depth.

\textbf{Visual Search.} Radiologists often revisit specific regions of an image when reassessing hypotheses, using localized inspection to verify findings. Inspired by this, we adapt a visual search mechanism~\cite{wu2024v} that iteratively focuses the proposer on salient regions. At iteration $t > 1$, given the top-$k$ candidate differences, the proposer predicts both candidate differences and normalized bounding boxes $\{x_1, y_1, x_2, y_2\} \in [0,1]^4$ indicating regions supporting each difference. We crop and recompose these regions into focused image grids and feed them to the proposer in the next iteration. This process improves local visual understanding, a key component of radiologist reasoning.

\section{Result}

In this section, we report the performance of \method{} on \benchmark{} and present careful ablation studies examining how each of the four methodological improvements contributes to the final performance.

\subsection{Overall Results}

Table~\ref{tab:main_result} summarizes the performance on \benchmark{}.

\textbf{\method{} achieves strong performance.} Our full system, \method{}, achieves 47.37\% top-1 accuracy—a dramatic improvement over the general-domain baseline VisDiff, which attains only 1.75\% on this highly challenging benchmark. This improvement is consistent across all difficulty levels: \method{} obtains 60.87\% (easy), 47.62\% (medium), and 23.08\% (hard), compared to VisDiff’s 4.35\%, 0.00\%, and 0.00\%, respectively.

\textbf{Performance can be improved with expert-written reports.} Since the proposer in \method{} uses CheXagent-generated radiology reports, we examine whether ground-truth expert-written reports provide additional benefits. We find a modest improvement—from 47.37\% to 50.88\% top-1 accuracy—indicating that while expert reports help, modern MLLMs (e.g., CheXagent) already generate radiology reports of sufficiently high fidelity for this task.

\textbf{Hard cases remain challenging.} Despite substantial gains over prior systems, the hard subset of \benchmark{} remains difficult: \method{} achieves only 23.08\% top-1 accuracy on these cases, compared to 47.37\% overall. These difficult groups capture subtle, clinically nuanced differences, highlighting opportunities for future research. Additional qualitative analyses are provided in the Appendix.

\begin{figure}[!tb]
  \centering
  \begin{subfigure}[!h]{\linewidth}
    \centering
    \includegraphics[width=\linewidth]{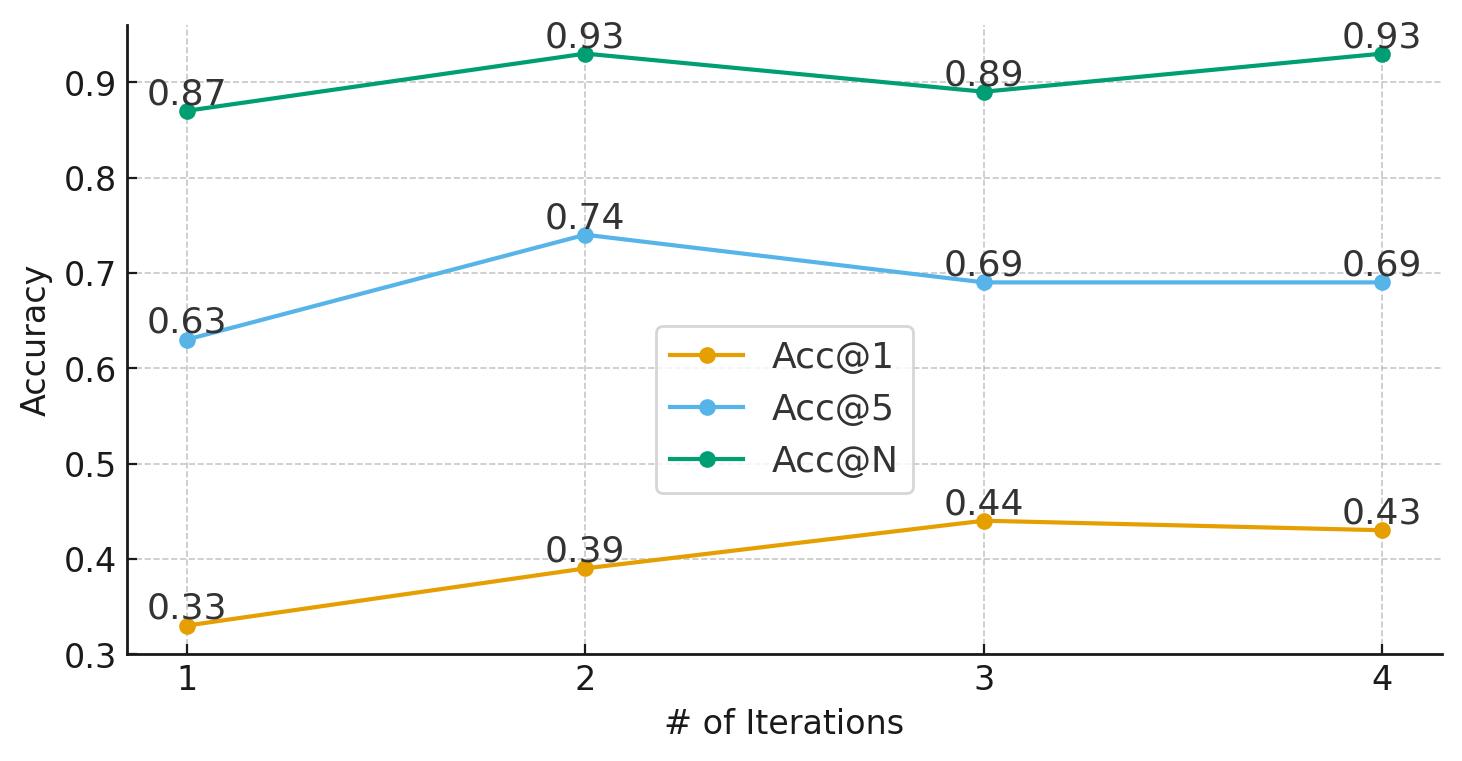}
    \label{fig:plot2}
  \end{subfigure}
  \vspace{-2.5em}
    \caption{\textbf{Ablation of iterative refinement rounds.} We find that iterative refinement improves performance, with the model plateauing around the third round.}
  \label{fig:threeplots}
  \vspace{-1.5em}
\end{figure}

\begin{figure*}[t]
  \centering
  \begin{minipage}[t]{0.48\linewidth}
    \centering
    \includegraphics[width=\linewidth, trim={0 1720 0 0}, clip]{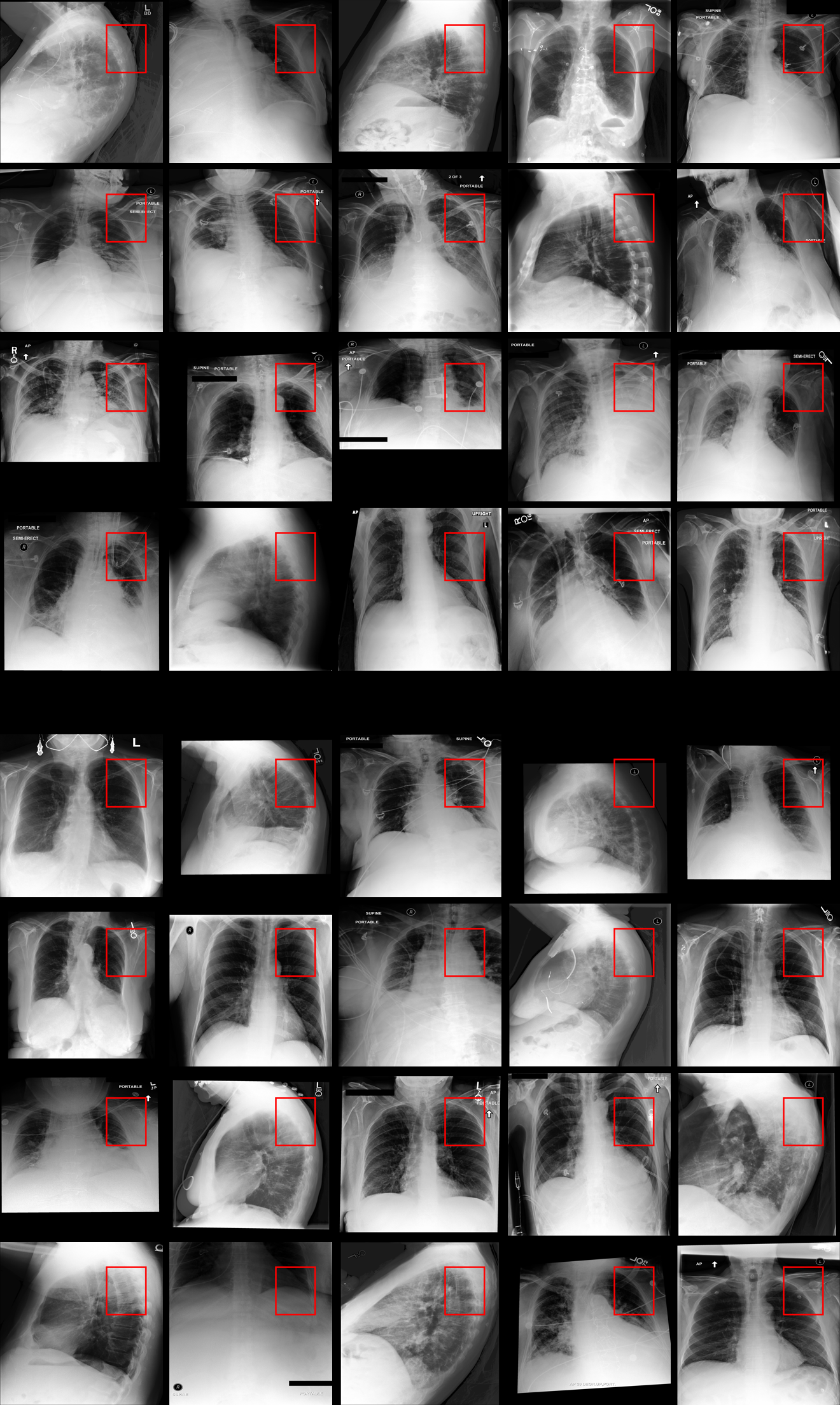}
    \caption*{(a) Pneumonia patients died in hospital.}
  \end{minipage}\hfill
  \begin{minipage}[t]{0.48\linewidth}
    \centering
    \includegraphics[width=\linewidth, trim={0 0 0 1720}, clip]{images/pneumonia_tc3_round1.png}
    \caption*{(b) Pneumonia patients survived in hospital.}
  \end{minipage}

  \vspace{4pt}
  \footnotesize
  \setlength{\tabcolsep}{6pt}
  \renewcommand{\arraystretch}{1.05}
  \begin{tabularx}{0.98\linewidth}{@{}lX@{}}
    \textbf{Category} & \textbf{Example difference} \\
    Lines/devices & Greater variety of tubes/catheters; repeated changes in chest tubes and right-sided PICC placement \\
    Parenchyma & More diffuse bilateral opacities and persistent infiltrates; higher pulmonary edema burden \\
    Pleural space & More bilateral pleural effusions (prevalence/size); small left pneumothorax events \\
    Volumes & Lower lung volumes and bibasilar atelectasis \\
  \end{tabularx}
\vspace{-0.5em}
  \caption{
  \textbf{Pneumonia non-survivors vs. pneumonia survivors.} 
Non-survivors show more extensive pulmonary disease and require more intensive interventions. 
The red box highlights dense device usage corresponding to the key difference: 
``more documented changes in thoracic catheters and lines.''
  }
  \label{fig:tte_combo_stack}
  \vspace{-1.5em}
\end{figure*}

\subsection{Ablations}

\textbf{Knowledge Injection.} Medical-domain models yield large gains over general-purpose systems. As shown in Table~\ref{tab:main_result}, the general-domain baseline (VisDiff) performs poorly on \benchmark{} (1.75\% top-1 accuracy), underscoring the difficulty of transferring generic visual reasoning to radiology. Introducing medical-specific components leads to immediate, substantial improvements: CheXagent captions raise accuracy to 9.65\%, and replacing the ranker with CheXzero increases performance to 28.95\%. These results demonstrate that medically pretrained VLMs supply essential domain grounding and radiological priors.

\textbf{Multimodal Reasoning.} Table~\ref{tab:main_result} shows that multimodal image–text inputs improve group-level radiology reasoning. Caption-only models miss subtle visual cues, whereas multimodal reasoning enables complementary use of textual descriptions and fine-grained spatial evidence. Our Joint Image \& Text variant achieves a 4-point top-1 accuracy gain (from 29.82\% to 33.33\%) over the captions-only version. This confirms that radiology captions and image features encode distinct, non-redundant signals essential for accurate clinical comparison.

\textbf{Iterative Refinement.} Iterative refinement consistently improves accuracy over single-pass reasoning (Table~\ref{tab:main_result}). Incorporating top-5 feedback boosts top-1 accuracy from 33.33\% to 43.86\% (an 11-point improvement). Gains are observed across all difficulty levels, with the most significant improvements occurring on hard cases where single-pass models often fail entirely. This demonstrates that iterative contextualization enables the system to refine hypotheses and suppress noise, mirroring how radiologists revisit and adjust interpretations over multiple passes. The ablation in Figure~\ref{fig:threeplots} further examines how iteration depth affects performance: accuracy peaks at the third iteration and then plateaus, likely due to reduced hypothesis diversity as more prior differences are recycled. Additional details are provided in the Appendix.

\textbf{Visual Search.} Combining iterative refinement with visual search yields our strongest overall performance (Table~\ref{tab:main_result}). The full \method{} system achieves 47.37\% top-1 accuracy and demonstrates strong performance across all difficulty levels, including 23.08\% top-1 accuracy on challenging cases where fine-grained spatial cues are crucial. Unlike pure iterative refinement, which only revisits textual hypotheses, visual search explicitly re-examines localized image regions by cropping high-saliency patches associated with previously identified differences. This dual refinement loop better mirrors radiologists’ practice of repeatedly inspecting specific regions of interest. Qualitatively, the model’s attention shifts toward clinically meaningful areas across iterations (e.g., gradually localizing a right pleural effusion), enhancing both localization and interpretability (see Appendix).

\begin{figure*}[t]
  \centering
  \begin{minipage}[t]{0.48\linewidth}
    \centering
    \includegraphics[width=\linewidth, trim={0 1720 0 0}, clip]{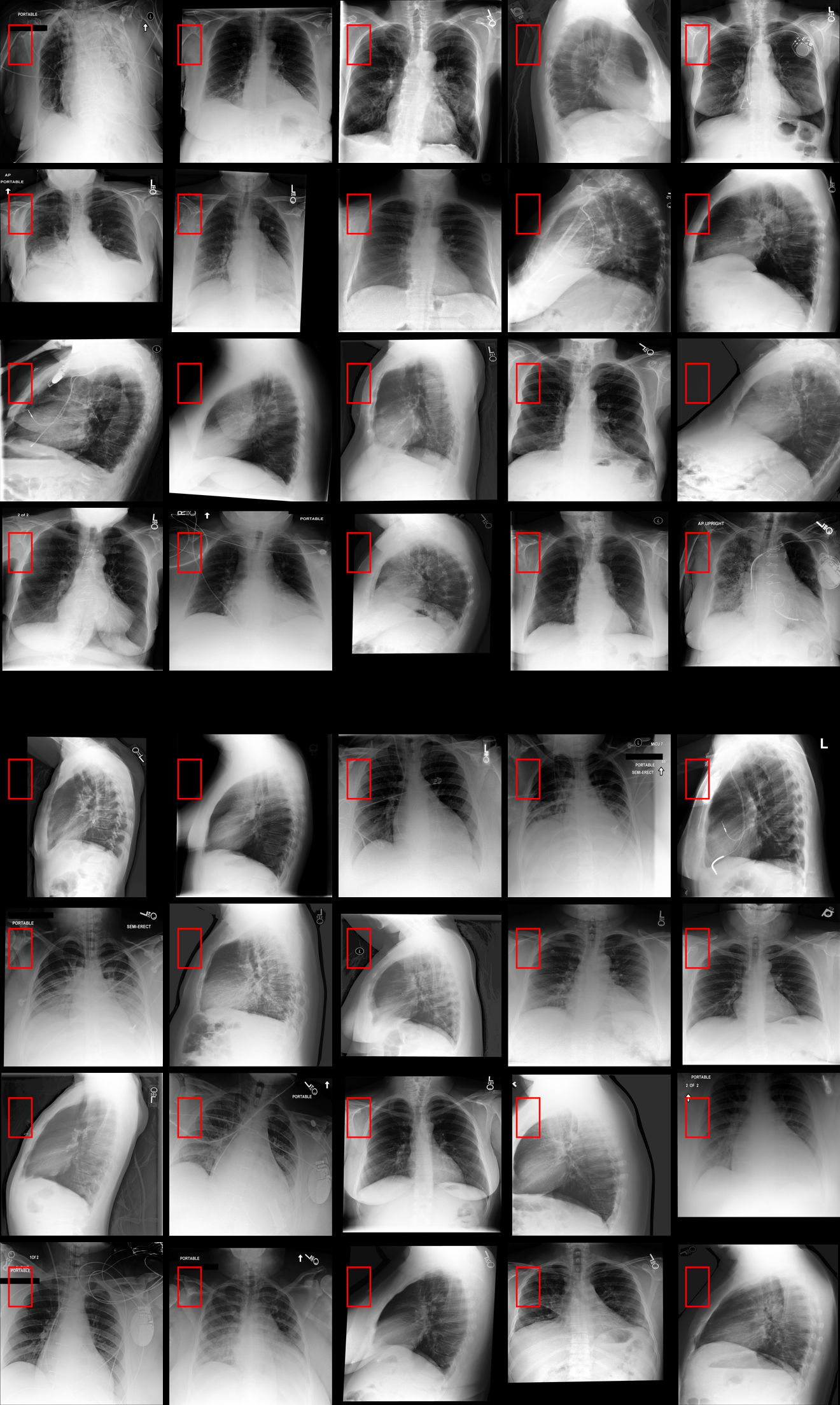}
    \caption*{(a) Older COVID-19 patients.}
  \end{minipage}\hfill
  \begin{minipage}[t]{0.48\linewidth}
    \centering
    \includegraphics[width=\linewidth, trim={0 0 0 1720}, clip]{images/old_vs_very_young_tc1_round2.png}
    \caption*{(b) Young COVID-19 patients.}
  \end{minipage}

  \vspace{4pt}
  \footnotesize
  \setlength{\tabcolsep}{6pt}
  \renewcommand{\arraystretch}{1.05}
  \begin{tabularx}{0.98\linewidth}{@{}p{0.48\linewidth}p{0.48\linewidth}@{}}
    \textbf{Higher in older COVID patients} & \textbf{Higher in young COVID patients} \\
    Hyperinflation and emphysema-like changes; flattened diaphragms &
      Normal mediastinal contours \\
    Chronic obstructive / COPD-like overexpansion &
      More presence of diffuse pulmonary opacifications (pulmonary edema) \\
    Pulmonary vascular congestion and interstitial edema &
      normal cardiomediastinal silhouette with no abnormalities \\
  \end{tabularx}
\vspace{-0.5em}
  \caption{\textbf{Older vs. younger COVID-19 patients.} Older patients show chronic structural changes, such as hyperinflation and emphysema-like features, while younger patients present more acute infectious opacities. The red crop focuses on regions with ``more frequent mention of hyperinflation and emphysema-like features.''}
  \label{fig:covid_age_combo}
  \vspace{-1.5em}
\end{figure*}

\section{Application}

Having developed \method{}, which achieves strong performance on \benchmark{}, we next apply it to downstream radiology tasks to answer clinically meaningful questions and enable both scientific discovery and model diagnosis. Notably, radiologists have verified the findings from this section and confirmed that they are both meaningful and clinically consistent.

\subsection{Survival Analysis of Pneumonia Patients}

\textbf{Research question.} Hospitalized pneumonia patients show wide variation in illness severity, and early identification of high-risk cases is crucial for timely intervention. Prior work has demonstrated that chest radiographs contain prognostic signals. \citet{kim2023deep} has shown that deep-learning models can predict 30-day mortality from chest radiographs, yet the specific imaging features associated with worse outcomes remain unclear. Therefore, we ask: which radiographic patterns distinguish pneumonia patients at elevated risk of early mortality?

\textbf{Experimental setup.}  
\label{6.1}
We apply \method{} to a \emph{time-to-event} analysis between two pneumonia cohorts where \seta{} includes patients who died during hospitalization and \setb{} comprises patients who survived or died at least one year later. To construct this dataset, we link MIMIC-CXR \cite{Johnson2019} with MIMIC-IV \cite{johnson2023mimic} clinical records, pair each radiograph with a mortality label, and filter pneumonia cases via ICD-9/10 codes (Appendix). We consider four mortality horizons: in-hospital, 30-day, 90-day, and 1-year or later.  
To reduce confounding, we stratify patients by age (ten bins) and gender, and sample equal numbers of surviving and deceased cases within each stratum.  

\textbf{Findings.} \method{} surfaces clinically meaningful differences between survivors and non-survivors (Figure~\ref{fig:tte_combo_stack}). Non-survivors show a greater variety and number of thoracic devices, e.g., ``PICC lines" and ``central venous catheters and endotracheal tubes projecting above the carina," reflecting greater intervention intensity and severe respiratory compromise. Beyond device burden, \method{} reveals more  extensive pulmonary disease in non-survivors, identifying ``extensive bilateral pulmonary opacities," ``pulmonary vascular congestion," ``enlarged cardiomediastinal silhouette with associated pleural effusions," and ``bibasilar atelectasis." These correspond to unresolved or progressive infection and align with prior findings that greater lung opacity correlates with increased pneumonia mortality \citep{kim2023deep}. Moreover, the nature of distinguishing features shifts with the prognostic window. For 30-day and 90-day mortality, device-related differences diminish, while parenchymal findings such as opacities and effusions become more dominant. The 90-day cohort also shows increased hyperinflation, suggesting chronic lung disease signals play a greater role in medium-term risk than acute invasive support. Overall, these demonstrate \method{}'s ability to deliver interpretable, radiologist-consistent imaging biomarkers for early risk assessment, enabling transparent model-assisted discovery in the clinical setting.

\vspace{-0.1em}
\subsection{Comparing Older vs. Younger COVID-19 Patients}

\vfill\null

\textbf{Research question.} Age is a major determinant of COVID-19 severity and clinical trajectory and may alter how infection appear on chest radiographs. Although age-dependent phenotypes are clinically important, the specific radiographic features that differ between older and younger COVID-19 patients have not been systematically characterized. This motivates the question: how does physiologic aging shape the imaging phenotype of COVID-19, and what features distinguish older from younger patients?

\textbf{Experimental setup.} We compare chest radiographs from two COVID-19 cohorts: older patients ($> 60$ years) and young patients ($\leq 40$ years) using the \method{} framework. Dataset construction, preprocessing pipeline, and clinical linkage follow Section \ref{6.1}. COVID-19 cases are identified via ICD-10 codes, and groups are gender-matched to reduce confounding. To assess robustness, we perform a bidirectional analysis, alternating which cohort is \seta{} and \setb{}. \method{} yields consistent differences in both configurations.

\begin{figure*}[!htbp]
  \centering
  \begin{minipage}[t]{0.48\linewidth}
    \centering
    \includegraphics[width=\linewidth, trim={0 1720 0 0}, clip]{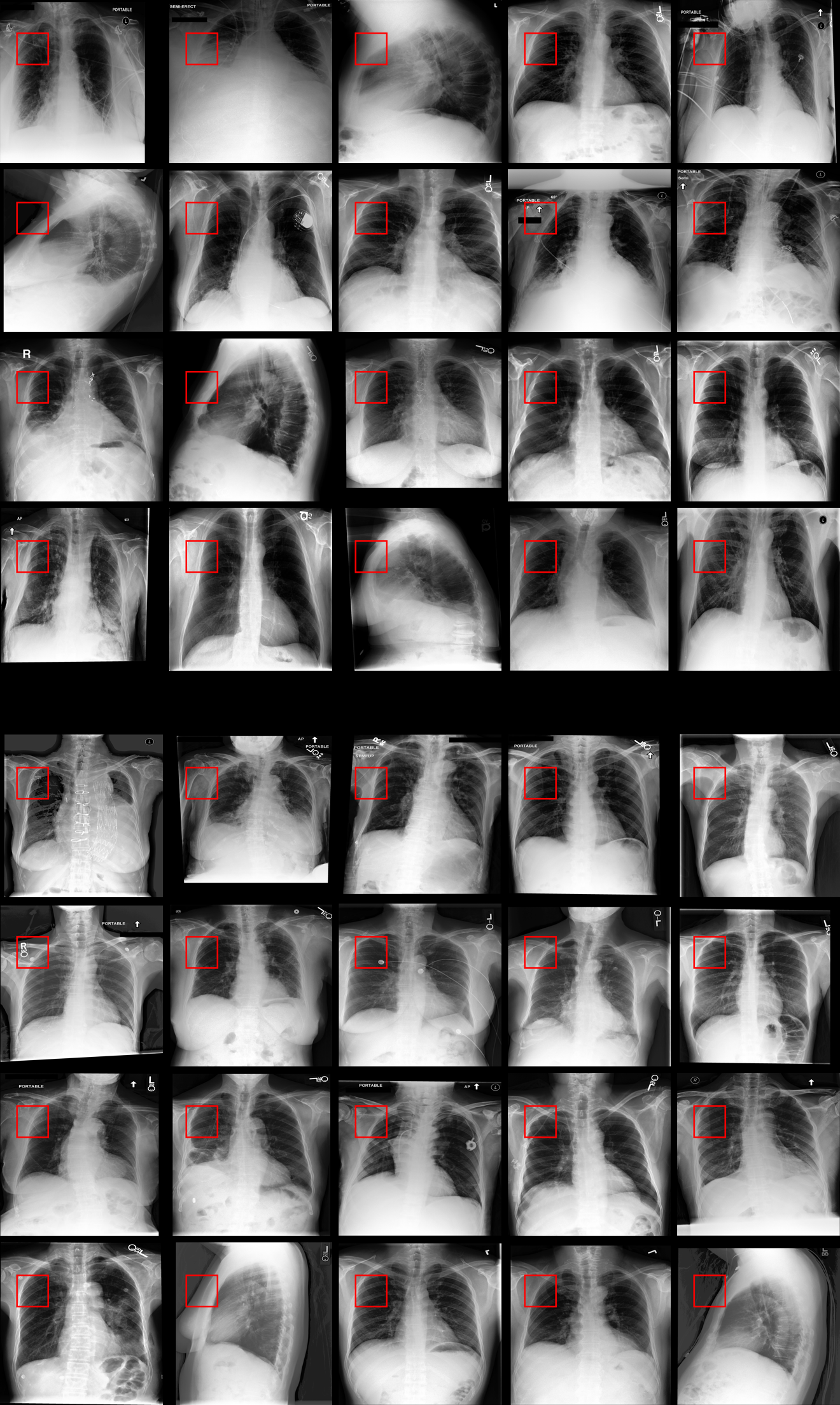}
    \caption*{(a) White cohort.}
  \end{minipage}\hfill
  \begin{minipage}[t]{0.48\linewidth}
    \centering
    \includegraphics[width=\linewidth, trim={0 0 0 1720}, clip]{images/race_tc1_round1.png}
    \caption*{(b) Asian cohort.}
  \end{minipage}

  \vspace{4pt}
  \footnotesize
  \setlength{\tabcolsep}{6pt}
  \renewcommand{\arraystretch}{1.05}
  \begin{tabularx}{0.98\linewidth}{@{}p{0.48\linewidth}p{0.48\linewidth}@{}}
    \textbf{Higher in White cohort} & \textbf{Higher in Asian cohort} \\
    Large hiatal hernia; hyperinflated lungs with flattened diaphragms &
      Small right apical pneumothorax \\
    More Port-A-Caths terminating at the cavoatrial junction &
      No focal consolidation or pneumothorax \\
    Small bilateral pleural effusions with atelectasis &
      Well-expanded lungs \\
    Large right pleural effusion with lobar collapse &
      Large right upper lobe mass with air bronchograms \\
    Endotracheal tube frequently 3.5\,cm above the carina &
      More reports of overall normal findings with some specific complications \\
  \end{tabularx}
\vspace{-0.5em}
  \caption{\textbf{Model classified White vs. Asian chest X-rays} The fine-tuned vision transformer underdiagnoses Asian patients relative to White patients, relying on spurious contextual cues rather than anatomy. The red box highlights the top difference ``more cases with the tip of vascular access devices (Port-A-Cath) terminating at the cavoatrial junction."}
  \label{fig:racial_combo}
  \vspace{-1.5em}
\end{figure*}

\textbf{Findings.} \method{} highlights clear and clinically coherent age-related distinctions (Figure~\ref{fig:covid_age_combo}). Older COVID-19 patients display features associated with chronic pulmonary remodeling, including ``lung hyperinflation resembling chronic obstructive pulmonary disease (COPD)", ``emphysema-like lucencies", and more pronounced ``vascular congestion and interstitial edema," suggesting that chronic background abnormalities modulate acute infection. In contrast, younger patients show fewer chronic structural changes, and ``more acute diffuse opacities", reflecting active infection and greater ventilatory reserve. Together, these demonstrate that \method{} uncovers interpretable age-dependent imaging phenotypes. By identifying how the same disease manifests differently across age groups, RaDiff offers clinically grounded insights that can support age-aware triage and enhance transparency in multimodal COVID-19 analysis.

\subsection{Discerning Racial Differences from Radiological Images}

\textbf{Research question.} Recent studies show that medical imaging models can predict race from chest X-rays with unexpectedly high accuracy \cite{gichoya2022ai, lotter2024acquisition}, despite clinicians being unable, and not trained, to infer race from these images. This raises significant concerns about what visual cues models exploit to make such predictions. Our goal is not to assert biological differences, but to uncover potential \emph{confounding factors} that may underlie race-predictive performance in medical vision models. This naturally leads to the question: what cues enable deep learning models to infer patient race from radiological images? 

\textbf{Experimental setup.} To investigate this separability, we apply \method{} to compare chest radiographs labeled as White (\seta{}) and Asian (\setb{}) patients, performing the analysis bi-directionally. We first fine-tune a DeiT-Small (patch16-224) Vision Transformer on an unstratified dataset of 15,000 chest X-rays (5k White, 5k Asian, 5k Black). After three epochs, the classifier reaches approximately 75\% validation accuracy. We then bootstrap a high-confidence subset (\(p_{\mathrm{race}} > 0.95\)) of White and Asian studies based on the classifier’s predictions. This subset forms the input to \method{}, which here functions as a model auditor.

\textbf{Findings.} \method{} primarily reveals procedural and contextual, rather than anatomical, differences. For example, the White cohort is associated with device-related details such as ``Port-A-Cath placements terminating at the cavoatrial junction" and ``endotracheal tubes positioned approximately 3.5 cm above the carina" (seen in Figure~\ref{fig:racial_combo}), patterns more reflective of institutional practice variation than patient physiology. More surprisingly, \method{} exposes strong asymmetries in reported normality. When White patients are \seta{}, top differences include abnormalities such as ``large hiatal hernia". Conversely, when Asian patients are \seta{}, \method{} instead highlights ``well-expanded lungs", ``no focal consolidation or pneumothorax", and ``overall normal lung findings." Three of the top five and seven of the top ten differences emphasize normality for Asian patients, whereas none do for White patients. This asymmetry mirrors the \emph{underdiagnosis bias} described by \citet{lotter2024acquisition}, where acquisition-related factors cause certain races to appear less abnormal, allowing models to learn spurious shortcuts and overlook pathology. These indicate that race-predictive performance in medical imaging models is largely driven by non-biological confounders. \method{} provides a principled framework for auditing such cues, informing dataset rebalancing, acquisition standardization, and fairness-aware model design, ultimately promoting medical AI systems that are clinically meaningful, equitable, and transparent.

\section{Conclusion}

We present \method{}, a multimodal agentic reasoning system designed to identify medically grounded differences between two radiological image sets. \method{} achieves strong performance on \benchmark{}, a newly developed, challenging benchmark for this task. Moreover, \method{} demonstrates practical utility across a wide variety of clinical applications, including survival analysis of pneumonia patients, comparisons between older and younger COVID-19 patients, and discerning racial differences in radiology images. Together, these contributions establish a foundation and toolset for building transparent, interpretable systems that reason over medical imaging differences and support scientific insight, clinical discovery, and the development of fairer medical AI.

{
    \small
    \bibliographystyle{ieeenat_fullname}
    \bibliography{main}

@String(ECCV= {Eur. Conf. Comput. Vis.})

@String(ECCV  = {ECCV})

@inproceedings{dunlap2024describing,
  title={Describing differences in image sets with natural language},
  author={Dunlap, Lisa and Zhang, Yuhui and Wang, Xiaohan and Zhong, Ruiqi and Darrell, Trevor and Steinhardt, Jacob and Gonzalez, Joseph E and Yeung-Levy, Serena},
  booktitle={Proceedings of the IEEE/CVF Conference on Computer Vision and Pattern Recognition},
  pages={24199--24208},
  year={2024}
}

@article{chen2024chexagent,
  title={CheXagent: Towards a Foundation Model for Chest X-Ray Interpretation},
  author={Chen, Zhihong and Varma, Maya and Delbrouck, Jean-Benoit and Paschali, Magdalini and Blankemeier, Louis and Van Veen, Dave and Valanarasu, Jeya Maria Jose and Youssef, Alaa and Cohen, Joseph Paul and Reis, Eduardo Pontes and others},
  journal={arXiv preprint arXiv:2401.12208},
  year={2024}
}

@article{tiu2022expert,
  title={Expert-level detection of pathologies from unannotated chest X-ray images via self-supervised learning},
  author={Tiu, Ekin and Talius, Ellie and Patel, Pujan and Langlotz, Curtis P and Ng, Andrew Y and Rajpurkar, Pranav},
  journal={Nature Biomedical Engineering},
   year={2022}
}

@inproceedings{wu2024v,
  title={V?: Guided visual search as a core mechanism in multimodal llms},
  author={Wu, Penghao and Xie, Saining},
  booktitle={Proceedings of the IEEE/CVF Conference on Computer Vision and Pattern Recognition},
  pages={13084--13094},
  year={2024}
}

@article{li2023mimicit,
  title={Mimic-it: Multi-modal in-context instruction tuning},
  author={Li, Bo and Zhang, Yuanhan and Chen, Liangyu and Wang, Jinghao and Pu, Fanyi and Yang, Jingkang and Li, Chunyuan and Liu, Ziwei},
  journal={arXiv preprint arXiv:2306.05425},
  year={2023}
}

@article{Johnson2019,
  title = {MIMIC-CXR,  a de-identified publicly available database of chest radiographs with free-text reports},
  volume = {6},
  ISSN = {2052-4463},
  url = {http://dx.doi.org/10.1038/s41597-019-0322-0},
  DOI = {10.1038/s41597-019-0322-0},
  number = {1},
  journal = {Scientific Data},
  publisher = {Springer Science and Business Media LLC},
  author = {Johnson,  Alistair E. W. and Pollard,  Tom J. and Berkowitz,  Seth J. and Greenbaum,  Nathaniel R. and Lungren,  Matthew P. and Deng,  Chih-ying and Mark,  Roger G. and Horng,  Steven},
  year = {2019},
  month = dec 
}

@article{dubois2023alpacafarm,
  title={Alpacafarm: A simulation framework for methods that learn from human feedback},
  author={Dubois, Yann and Li, Xuechen and Taori, Rohan and Zhang, Tianyi and Gulrajani, Ishaan and Ba, Jimmy and Guestrin, Carlos and Liang, Percy and Hashimoto, Tatsunori B},
  journal={arXiv preprint arXiv:2305.14387},
  year={2023}
}

@article{gichoya2022ai,
  title={AI recognition of patient race in medical imaging: a modelling study},
  author={Gichoya, Judy Wawira and Banerjee, Imon and Bhimireddy, Ananth Reddy and Burns, John L and Celi, Leo Anthony and Chen, Li-Ching and Correa, Ramon and Dullerud, Natalie and Ghassemi, Marzyeh and Huang, Shih-Cheng and others},
  journal={The Lancet Digital Health},
  volume={4},
  number={6},
  pages={e406--e414},
  year={2022},
  publisher={Elsevier}
}

@article{li2023blip,
  title={Blip-2: Bootstrapping language-image pre-training with frozen image encoders and large language models},
  author={Li, Junnan and Li, Dongxu and Savarese, Silvio and Hoi, Steven},
  journal={arXiv preprint arXiv:2301.12597},
  year={2023}
}

@inproceedings{radford2021learning,
  title={Learning transferable visual models from natural language supervision},
  author={Radford, Alec and Kim, Jong Wook and Hallacy, Chris and Ramesh, Aditya and Goh, Gabriel and Agarwal, Sandhini and Sastry, Girish and Askell, Amanda and Mishkin, Pamela and Clark, Jack and others},
  booktitle={ICML},
  year={2021}
}

@article{johnson2023mimic,
  title={MIMIC-IV, a freely accessible electronic health record dataset},
  author={Johnson, Alistair EW and Bulgarelli, Lucas and Shen, Lu and Gayles, Alvin and Shammout, Ayad and Horng, Steven and Pollard, Tom J and Hao, Sicheng and Moody, Benjamin and Gow, Brian and others},
  journal={Scientific data},
  volume={10},
  number={1},
  pages={1},
  year={2023},
  publisher={Nature Publishing Group UK London}
}

@article{okoye2022computed,
  title={Computed tomography findings and prognosis in older COVID-19 patients},
  author={Okoye, Chukwuma and Finamore, Panaiotis and Bellelli, Giuseppe and Coin, Alessandra and Del Signore, Susanna and Fumagalli, Stefano and Gareri, Pietro and Malara, Alba and Mossello, Enrico and Trevisan, Caterina and others},
  journal={BMC geriatrics},
  volume={22},
  number={1},
  pages={166},
  year={2022},
  publisher={Springer}
}

@inproceedings{zhai2023sigmoid,
  title={Sigmoid loss for language image pre-training},
  author={Zhai, Xiaohua and Mustafa, Basil and Kolesnikov, Alexander and Beyer, Lucas},
  booktitle={Proceedings of the IEEE/CVF international conference on computer vision},
  pages={11975--11986},
  year={2023}
}

@inproceedings{girdhar2023imagebind,
  title={Imagebind: One embedding space to bind them all},
  author={Girdhar, Rohit and El-Nouby, Alaaeldin and Liu, Zhuang and Singh, Mannat and Alwala, Kalyan Vasudev and Joulin, Armand and Misra, Ishan},
  booktitle={Proceedings of the IEEE/CVF conference on computer vision and pattern recognition},
  pages={15180--15190},
  year={2023}
}

@inproceedings{wang2024videoagent,
  title={Videoagent: Long-form video understanding with large language model as agent},
  author={Wang, Xiaohan and Zhang, Yuhui and Zohar, Orr and Yeung-Levy, Serena},
  booktitle={European Conference on Computer Vision},
  pages={58--76},
  year={2024},
  organization={Springer}
}

@article{shen2023hugginggpt,
  title={Hugginggpt: Solving ai tasks with chatgpt and its friends in hugging face},
  author={Shen, Yongliang and Song, Kaitao and Tan, Xu and Li, Dongsheng and Lu, Weiming and Zhuang, Yueting},
  journal={Advances in Neural Information Processing Systems},
  volume={36},
  pages={38154--38180},
  year={2023}
}

@inproceedings{gupta2023visual,
  title={Visual programming: Compositional visual reasoning without training},
  author={Gupta, Tanmay and Kembhavi, Aniruddha},
  booktitle={Proceedings of the IEEE/CVF conference on computer vision and pattern recognition},
  pages={14953--14962},
  year={2023}
}

@inproceedings{suris2023vipergpt,
  title={Vipergpt: Visual inference via python execution for reasoning},
  author={Sur{\'\i}s, D{\'\i}dac and Menon, Sachit and Vondrick, Carl},
  booktitle={Proceedings of the IEEE/CVF international conference on computer vision},
  pages={11888--11898},
  year={2023}
}

@article{robertson2009probabilistic,
  title={The probabilistic relevance framework: BM25 and beyond},
  author={Robertson, Stephen and Zaragoza, Hugo and others},
  journal={Foundations and Trends{\textregistered} in Information Retrieval},
  volume={3},
  number={4},
  pages={333--389},
  year={2009},
  publisher={Now Publishers, Inc.}
}

@inproceedings{llava1.5,
  title={Visual instruction tuning},
  author={Liu, Haotian and Li, Chunyuan and Wu, Qingyang and Lee, Yong Jae},
  booktitle={NeurIPS},
  year={2023}
}

@article{gpt4,
      title={GPT-4 Technical Report}, 
      author={OpenAI},
      year={2023},
      journal={arXiv preprint arXiv:2303.08774}
}

@inproceedings{
wang2022multigranularity,
title={Multi-Granularity Cross-modal Alignment for Generalized Medical Visual Representation Learning},
author={Fuying Wang and Yuyin Zhou and Shujun Wang and Varut Vardhanabhuti and Lequan Yu},
booktitle={Advances in Neural Information Processing Systems},
editor={Alice H. Oh and Alekh Agarwal and Danielle Belgrave and Kyunghyun Cho},
year={2022},
url={https://openreview.net/forum?id=Yul402KcD5d}
}

@misc{zhang2022contrastivelearningmedicalvisual,
      title={Contrastive Learning of Medical Visual Representations from Paired Images and Text}, 
      author={Yuhao Zhang and Hang Jiang and Yasuhide Miura and Christopher D. Manning and Curtis P. Langlotz},
      year={2022},
      eprint={2010.00747},
      archivePrefix={arXiv},
      primaryClass={cs.CV},
      url={https://arxiv.org/abs/2010.00747}, 
}

@inproceedings{huang2021gloria,
  title={GLoRIA: A Multimodal Global-Local Representation Learning Framework for Label-Efficient Medical Image Recognition},
  author={Huang, Shih-Cheng and Shen, Liyue and Lungren, Matthew P and Yeung, Serena},
  booktitle={Proceedings of the IEEE/CVF International Conference on Computer Vision},
  pages={3942--3951},
  year={2021}
}

@inbook{Boecking_2022,
   title={Making the Most of Text Semantics to Improve Biomedical Vision–Language Processing},
   ISBN={9783031200595},
   ISSN={1611-3349},
   url={http://dx.doi.org/10.1007/978-3-031-20059-5_1},
   DOI={10.1007/978-3-031-20059-5_1},
   booktitle={Computer Vision – ECCV 2022},
   publisher={Springer Nature Switzerland},
   author={Boecking, Benedikt and Usuyama, Naoto and Bannur, Shruthi and Castro, Daniel C. and Schwaighofer, Anton and Hyland, Stephanie and Wetscherek, Maria and Naumann, Tristan and Nori, Aditya and Alvarez-Valle, Javier and Poon, Hoifung and Oktay, Ozan},
   year={2022},
   pages={1–21} }

@misc{sellergren2025medgemmatechnicalreport,
      title={MedGemma Technical Report}, 
      author={Andrew Sellergren and Sahar Kazemzadeh and Tiam Jaroensri and Atilla Kiraly and Madeleine Traverse and Timo Kohlberger and Shawn Xu and Fayaz Jamil and Cían Hughes and Charles Lau and Justin Chen and Fereshteh Mahvar and Liron Yatziv and Tiffany Chen and Bram Sterling and Stefanie Anna Baby and Susanna Maria Baby and Jeremy Lai and Samuel Schmidgall and Lu Yang and Kejia Chen and Per Bjornsson and Shashir Reddy and Ryan Brush and Kenneth Philbrick and Mercy Asiedu and Ines Mezerreg and Howard Hu and Howard Yang and Richa Tiwari and Sunny Jansen and Preeti Singh and Yun Liu and Shekoofeh Azizi and Aishwarya Kamath and Johan Ferret and Shreya Pathak and Nino Vieillard and Ramona Merhej and Sarah Perrin and Tatiana Matejovicova and Alexandre Ramé and Morgane Riviere and Louis Rouillard and Thomas Mesnard and Geoffrey Cideron and Jean-bastien Grill and Sabela Ramos and Edouard Yvinec and Michelle Casbon and Elena Buchatskaya and Jean-Baptiste Alayrac and Dmitry Lepikhin and Vlad Feinberg and Sebastian Borgeaud and Alek Andreev and Cassidy Hardin and Robert Dadashi and Léonard Hussenot and Armand Joulin and Olivier Bachem and Yossi Matias and Katherine Chou and Avinatan Hassidim and Kavi Goel and Clement Farabet and Joelle Barral and Tris Warkentin and Jonathon Shlens and David Fleet and Victor Cotruta and Omar Sanseviero and Gus Martins and Phoebe Kirk and Anand Rao and Shravya Shetty and David F. Steiner and Can Kirmizibayrak and Rory Pilgrim and Daniel Golden and Lin Yang},
      year={2025},
      eprint={2507.05201},
      archivePrefix={arXiv},
      primaryClass={cs.AI},
      url={https://arxiv.org/abs/2507.05201}, 
}

@misc{bannur2024maira2groundedradiologyreport,
      title={MAIRA-2: Grounded Radiology Report Generation}, 
      author={Shruthi Bannur and Kenza Bouzid and Daniel C. Castro and Anton Schwaighofer and Anja Thieme and Sam Bond-Taylor and Maximilian Ilse and Fernando Pérez-García and Valentina Salvatelli and Harshita Sharma and Felix Meissen and Mercy Ranjit and Shaury Srivastav and Julia Gong and Noel C. F. Codella and Fabian Falck and Ozan Oktay and Matthew P. Lungren and Maria Teodora Wetscherek and Javier Alvarez-Valle and Stephanie L. Hyland},
      year={2024},
      eprint={2406.04449},
      archivePrefix={arXiv},
      primaryClass={cs.CL},
      url={https://arxiv.org/abs/2406.04449}, 
}

@article{kim2023deep,
  title={A deep learning model using chest radiographs for prediction of 30-day mortality in patients with community-acquired pneumonia: development and external validation},
  author={Kim, Changi and Hwang, Eui Jin and Choi, Ye Ra and Choi, Hyewon and Goo, Jin Mo and Kim, Yisak and Choi, Jinwook and Park, Chang Min},
  journal={American Journal of Roentgenology},
  volume={221},
  number={5},
  pages={586--598},
  year={2023},
  publisher={American Roentgen Ray Society}
}

@article{lotter2024acquisition,
  title={Acquisition parameters influence AI recognition of race in chest x-rays and mitigating these factors reduces underdiagnosis bias},
  author={Lotter, William},
  journal={Nature communications},
  volume={15},
  number={1},
  pages={7465},
  year={2024},
  publisher={Nature Publishing Group UK London}
}
}

\clearpage
\maketitlesupplementary
\appendix

\section*{Acknowledgments} 

S.Y. is a Chan Zuckerberg Biohub — San Francisco Investigator.

\section*{Reproducibility Statement}
We provide code implementations of \method{} and \benchmark{} at \url{https://github.com/yuhui-zh15/RadDiff}.

\section*{Limitations}
While \method{} demonstrates strong performance across diverse radiological difference identification tasks, several limitations remain. First, the framework still has room for improvement on particularly challenging subsets, where even small inconsistencies in cropping or ranking may propagate through iterative refinement. Second, \method{} should be used with a human-in-the-loop, especially for high-stakes applications involving prognosis or outcome prediction; our system is designed to surface candidate differences, not to replace expert review.

\section*{Table of Contents}
In this supplementary material, we provide additional information of \benchmark{}, \method{}, results, and applications.

In Appendix A, we present a breakdown of \benchmark{}, including the details of the benchmark creation process, the evaluator prompt, and examples from \benchmark{}.

In Appendix B, we describe the prompts used for multimodal reasoning, iterative refinement, and visual search.

In Appendix C, we provide additional qualitative analyses, a detailed case study demonstrating \method{} difference discovery, and further ablations exploring experimental design choices.

In Appendix D, we provide extended details on the application-level experiments, including setup, and supplementary qualitative results.

\section{Supplementary Section 3}
In this section, we provide additional details of Section 3 in the main paper.

\subsection{Differences between Paired Radiology Image Sets}
We provide all Easy/Medium/Hard subset differences for the paired radiology image sets in \benchmark{} in Table~\ref{tab:diff_examples}.

\begin{table}[t]
\tiny
\centering
\renewcommand{\arraystretch}{0.9}
\setlength{\tabcolsep}{1.5pt}

\begin{tabular}{p{3.3cm}|p{3.3cm}}
\toprule
\textbf{Set A} & \textbf{Set B} \\
\midrule

\multicolumn{2}{c}{\textit{\textbf{Easy (23 examples)}}} \\
\midrule
right subclavian central venous catheter present & no subclavian central venous catheter observe \\
previously note pulmonary edema resolve & moderate pulmonary edema with slightly improve aeration \\
mild to moderate cardiomegaly & moderate cardiomegaly, mildly stable \\
NG tube in bronchus & NG tube in stomach \\
unstable cardiomegaly with pulmonary edema & stable cardiomegaly, no edema \\
abnormal chest radiograph & Normal chest radiograph \\
Left apical pleural tube in place & no pleural tube \\
enteric tube terminate below diaphragm & no enteric tube \\
poc catheter tip in the low SVC & poc catheter not visualize \\
heart size enlarge & Normal heart size \\
single view chest radiograph & PA and lateral chest radiograph \\
enlarged cardiac silhouette & Normal cardiac silhouette \\
patchy middle lobe opacity & Clear middle lobe \\
nasogastric tube remove & nasogastric tube present \\
hyperinflated lung w/ flattened diaphragms & clear lung \\
PICC terminate in low SVC & PICC absent or not in low SVC \\
mild cardiomegaly & significant cardiomegaly \\
opacity concern for pneumonia & no evidence of pneumonia \\
clear lung w/o consolidation/effusion/pneumothorax & right low lobe pneumonia \\
interstitial abnormality w/ vascular congestion & no interstitial abnormality or congestion \\
right lung residual patchy opacity & clear right lung \\
hyperinflated lung & Normal lung inflation \\
moderate–large right pneumothorax & no pneumothorax \\

\midrule
\multicolumn{2}{c}{\textit{\textbf{Medium (21 examples)}}} \\
\midrule
moderate right pleural effusion & no pleural effusion \\
Hypoinflated lungs w/ perihilar opacity & lungs well inflated and clear \\
bilateral small pleural effusion & no pleural effusion or pneumothorax \\
small–moderate left pneumothorax & no pneumothorax \\
bilateral pneumothorax & no pneumothorax \\
Clear lung & diffuse interstitial opacity \\
right middle lobe pneumonia & no pneumonia \\
subtle opacity left lung base & clear lung base \\
moderate–severe cardiomegaly & Normal heart size \\
lungs well inflated, clear & bibasal interstitial opacity \\
heart normal size appearance & heart mildly–moderately enlarged \\
new opacity left mid/lower lung & no new opacity \\
dense RUL consolidation & no consolidation \\
high sensitivity for pneumothorax & low sensitivity (supine) \\
lungs mostly clear & bibasilar opacity, lung mass \\
Rightward mediastinal shift & no shift \\
right apical opacity & no apical opacity \\
small bilateral pleural effusion & no effusion \\
sign of tuberculosis infection & no evidence of tuberculosis \\
moderate edema + effusion & minimal edema, no effusion \\
low lung volume + bibasilar opacity & normal lung volume, clear lung \\

\midrule
\multicolumn{2}{c}{\textit{\textbf{Hard (13 examples)}}} \\
\midrule
displace rib fracture & no displace rib fracture \\
stable airspace consolidation & worsen airspace consolidation \\
confluent left perihilar opacity & clear perihilar region \\
elevated pulmonary venous pressure & Normal venous pressure \\
lo lung volume & Normal lung volume \\
esophageal perforation & no perforation \\
heart size be normal & silhouette remain enlarged \\
pulmonary nodule & no pulmonary nodule \\
clear basal parenchyma & basal atelectasis \\
worsen retrocardiac opacification & no significant change \\
multilevel spinal degenerative change & Normal spinal structure \\
heart mildly enlarged/unchanged & heart not enlarged \\
moderate cardiomegaly & Normal silhouette \\
\bottomrule
\end{tabular}
\caption{Radiologist-validated \seta{} (Set A) and \setb{} (Set B) differences grouped by difficulty.}
\label{tab:diff_examples}
\end{table}

\subsection{Prompts for RadDiffBench construction}
We provide the prompts used for hypothetical difference proposal from reports, difference de-duplication, and report-based classification in Figures \ref{fig:difference_proposal},  \ref{fig:difference_dedup}, \ref{fig:difference_classification}.

\begin{figure*}[t]
\centering

% Header bar
\fcolorbox{black}{black}{
  \parbox{0.97\textwidth}{
    \vspace{2pt}
    \color{white}\large\bfseries Hypothetical Differences Proposal Prompt
    \vspace{2pt}
  }
}

\vspace{-1pt}

% Main prompt box
\fcolorbox{black}{gray!5}{
\parbox{0.97\textwidth}{
\small\ttfamily
\begingroup
\obeylines

List all hypothetical potential differences between sets of chest x-ray radiology scans.

These could include but not limited to variations in tissue density, presence of abnormalities such as tumors,
lesions, or fractures, and any noticeable changes in anatomical structures. 

Give me exactly \{num\_differences\} differences in the format of A vs B in a JSON file. 

Store condition A and B in seperate fields in the JSON. The JSON format should be of the following:

[
\{\{ "condition\_A": "*insert condition A*", "condition\_B": "*insert condition B*" \}\},
  ...
]

Ensure these distinctions reflect the detailed nuances characteristic of radiology reports. 

They should not be broad classification differences but rather subtle, intricate variations. 

Here are sample radiology reports to help you:

\{sample\_reports\}

\endgroup
}
}

\caption{Prompt used for Hypothetical Differences Proposal}
\label{fig:difference_proposal}
\end{figure*}

\begin{figure*}[t]
\centering

% Header bar
\fcolorbox{black}{black}{
  \parbox{0.97\textwidth}{
    \vspace{2pt}
    \color{white}\large\bfseries Proposal De-duplication Prompt
    \vspace{2pt}
  }
}

\vspace{-1pt}

% Main prompt box
\fcolorbox{black}{gray!5}{
\parbox{0.97\textwidth}{
\small\ttfamily
\begingroup
\obeylines

Below are hypothetical differences between chest X ray. For the below set of differences, remove any differences that are semantically and medically similar to each other. 

Please be sure to tell me which differences were removed and explain your reasoning.

\{differences\}

Return the final differences, with duplicates removed, as a JSON in the following format:

\{\{
differences: [
\{\{
    "condition\_A": "",
    "condition\_B": "",
\}\},
...
 ]
\}\}
            
\endgroup
}
}

\caption{Prompt used for Hypothetical Differences De-duplication}
\label{fig:difference_dedup}
\end{figure*}

\begin{figure*}[t]
\centering

% Header bar
\fcolorbox{black}{black}{
  \parbox{0.97\textwidth}{
    \vspace{2pt}
    \color{white}\large\bfseries Radiology Reports Classification Prompt
    \vspace{2pt}
  }
}

\vspace{-1pt}

% Main prompt box
\fcolorbox{black}{gray!5}{
\parbox{0.97\textwidth}{
\small\ttfamily
\begingroup
\obeylines

We have the following condition of the format A vs B respectively: \{difference\}.

Given the following \{len(reports)\} radiology reports, group each report into either having condition A or B or neither. 

Classify each report into only one group exactly. Do not place a report in multiple groups.

Provide reasoning and direct evidence in quotes from the report to justify each grouping. 

Put the final output in a JSON with the following format:

\{\{

"group A": [
    \{\{
          "report\_index": "",
          "reasoning": "",
          "direct\_evidence": "",
    \}\},
    ...
],

"group B": [
    \{\{
          "report\_index": "",
          "reasoning": "",
          "direct\_evidence": "",
    \}\},
    ...
],

"neither": [
    \{\{
          "report\_index": "",
          "reasoning": "",
          "direct\_evidence": "",
    \}\},
    ...
]

\}\}

Please make sure to classify ALL the reports shown below:

\{reports\}
      
\endgroup
}
}

\caption{Prompt used for Radiology Reports Classification}
\label{fig:difference_classification}
\end{figure*}

\subsection{Evaluator Prompt}
We provide the prompt used by GPT-4.1-nano during evaluation in Figure \ref{fig:eval_prompt}.

\begin{figure*}[t]
\centering

% Header bar
\fcolorbox{black}{black}{
  \parbox{0.97\textwidth}{
    \vspace{2pt}
    \color{white}\large\bfseries RadDiff Evaluator Prompt 
    \vspace{2pt}
  }
}

\vspace{-1pt}

% Main prompt box
\fcolorbox{black}{gray!5}{
\parbox{0.97\textwidth}{
\small\ttfamily
\begingroup
\obeylines

I am a machine learning researcher working on identifying differences between two groups of radiology images. My goal is to determine how well a given prediction corresponds to the findings or conditions that are more commonly present in Group A compared to Group B.

You will receive descriptions of Group A and Group B, along with a prediction. 
    
Your task: 

Evaluate whether the prediction is more aligned with Group A or Group B, using the following scoring system:

2: Fully aligned with Group A.

1: Partially aligned with Group A (i.e., the prediction is closer to Group A than Group B but represents a broader or narrower concept).

0: Not aligned with Group A (i.e., more aligned with Group B or represents a completely different concept).

Reference Example 1:

Group A: "Left-sided opacity" and Group B: "Right-sided opacity"

Prediction: "Left-sided opacity" → Score: 2 (fully aligned with Group A)

Prediction: "Left lung consolidation" → Score: 2 (fully aligned with Group A)

Prediction: "Unilateral lung opacity" → Score: 1 (broader but closer to Group A)

Prediction: "Right-sided opacity" → Score: 0 (aligned with Group B)

Reference Example 2:

Group A: "Pleural effusion" and Group B: "No pleural effusion"

Prediction: "Pleural effusion" → Score: 2 (fully aligned with Group A)

Prediction: "Fluid in the pleural space" → Score: 2 (fully aligned with Group A)

Prediction: "Increased fluid in the chest cavity" → Score: 1 (broader but closer to Group A)

Prediction: "Normal lungs" → Score: 0 (aligned with Group B)

Now, analyze the following using similar reasoning from the above examples as a guide. 

Group A: \{gt\_a\}

Group B: \{gt\_b\}

Prediction: \{hypothesis\}

Please respond with 2, 1, or 0, based on the alignment of the prediction with Group A.

\endgroup
}
}

\caption{Prompt used for LLM-based evaluator scoring candidate differences between Set~A \seta{} and Set~B \setb{}.}
\label{fig:eval_prompt}
\end{figure*}

\subsection{Examples for RadDiffBench}
\begin{figure*}[!htbp]
    \centering

    % -------- Easy 1 --------
    \begin{minipage}[t]{0.48\linewidth}
        \centering
        \includegraphics[width=\linewidth, trim={0 1720 0 0}, clip]{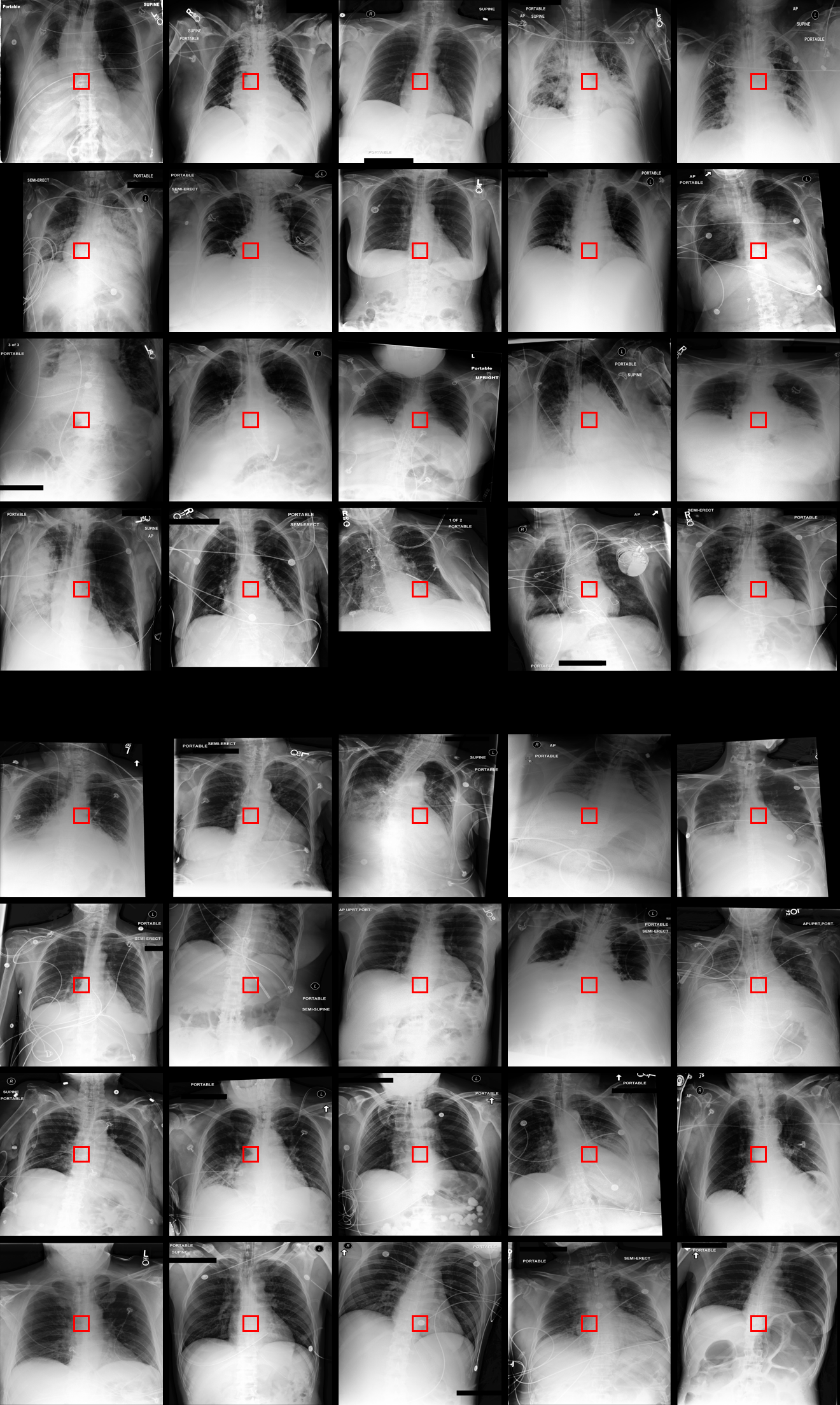}
        \caption*{\seta{}: NG tube in bronchus.}
    \end{minipage}\hfill
    \begin{minipage}[t]{0.48\linewidth}
        \centering
        \includegraphics[width=\linewidth, trim={0 0 0 1720}, clip]{images/easy_1_carina.png}
        \caption*{\setb{}: NG tube in stomach.}
    \end{minipage}

    \vspace{6pt}

    % -------- Easy 2 --------
    \begin{minipage}[t]{0.48\linewidth}
        \centering
        \includegraphics[width=\linewidth, trim={0 1720 0 0}, clip]{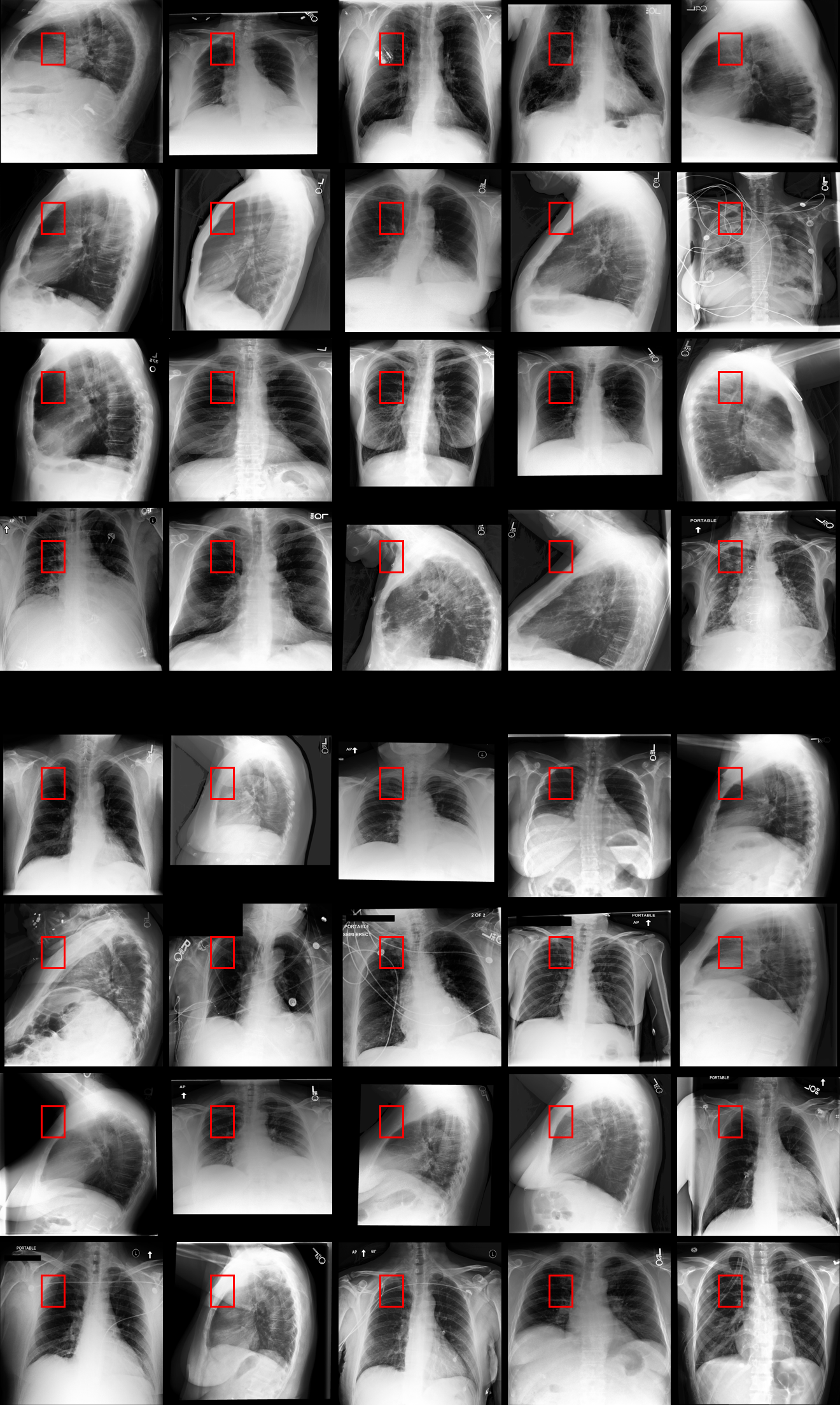}
        \caption*{\seta{}: Hyperinflated lung with flattening of the hemidiaphragm.}
    \end{minipage}\hfill
    \begin{minipage}[t]{0.48\linewidth}
        \centering
        \includegraphics[width=\linewidth, trim={0 0 0 1720}, clip]{images/easy_2.png}
        \caption*{\setb{}: Clear lung.}
    \end{minipage}

    \vspace{6pt}

    % -------- Easy 3 --------
    \begin{minipage}[t]{0.48\linewidth}
        \centering
        % top / middle crop: patchy middle lobe opacity
        \includegraphics[width=\linewidth, trim={0 1720 0 0}, clip]{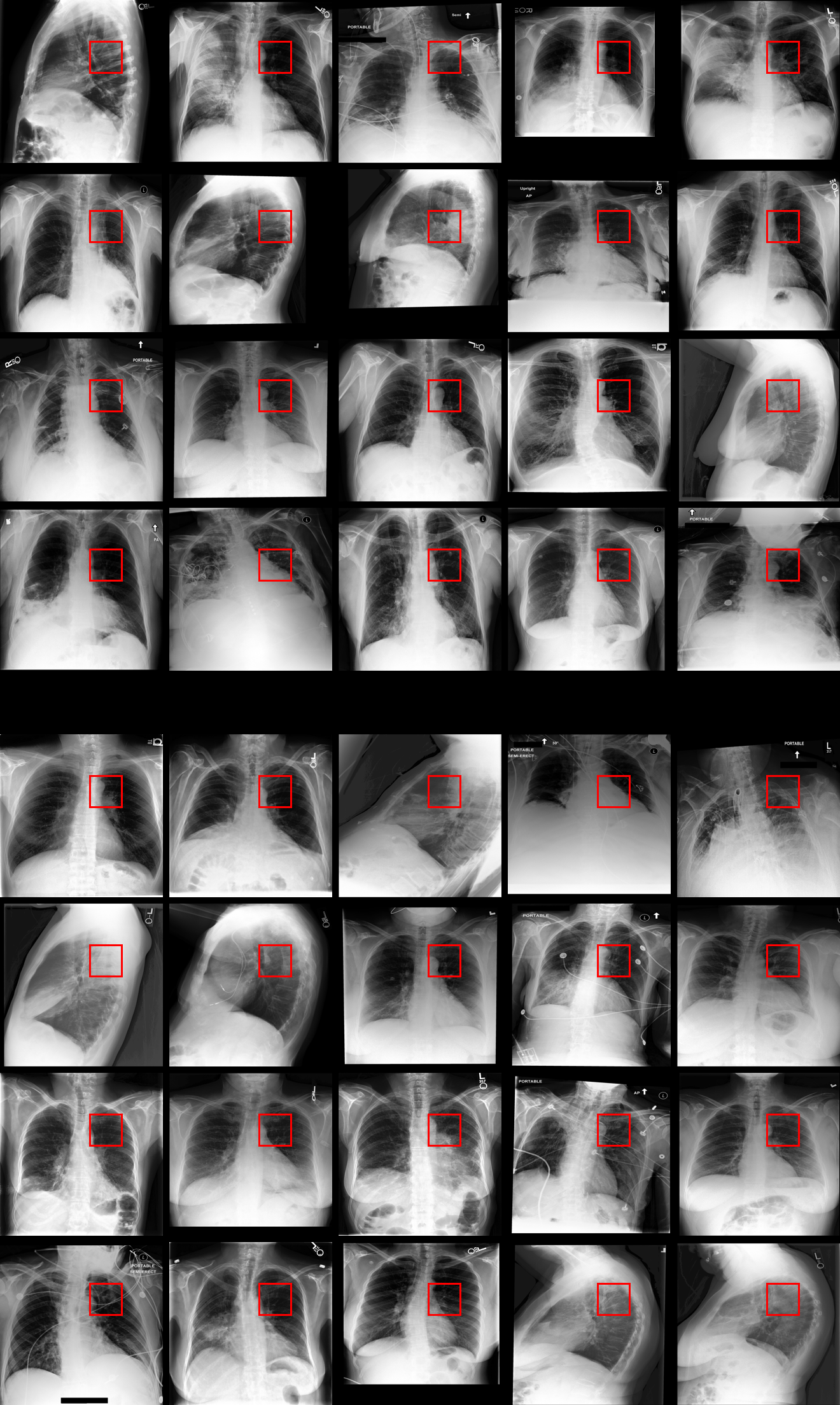}
        \caption*{\seta{}: Patchy middle lobe opacity.}
    \end{minipage}\hfill
    \begin{minipage}[t]{0.48\linewidth}
        \centering
        % adjust trim here if you want a slightly different region for the "clear" comparison
        \includegraphics[width=\linewidth, trim={0 0 0 1720}, clip]{images/easy_3.png}
        \caption*{\setb{}: Clear middle lobe.}
    \end{minipage}

    \caption{\textbf{Easy Examples.} \method{} localizes salient regions and surfaces clinically meaningful cohort-level differences. 
    Top row: carina region, producing predictions such as ``Higher frequency of endotracheal tubes located above the carina in Group~A.'' 
    Middle row: lung hyperinflation, producing ``More instances of hyperinflated lungs without focal consolidation or effusion in Group~A.'' 
    Bottom row: \method{} proposing differences such as ``More consolidation and opacity patterns suggestive of pneumonia in Group~A.''}
    \label{fig:easy_examples}
\end{figure*}

\begin{figure*}[!htbp]
    \centering

    % -------- Medium 1 --------
    \begin{minipage}[t]{0.48\linewidth}
        \centering
        % pneumonia in right middle lobe
        \includegraphics[width=\linewidth, trim={0 1720 0 0}, clip]{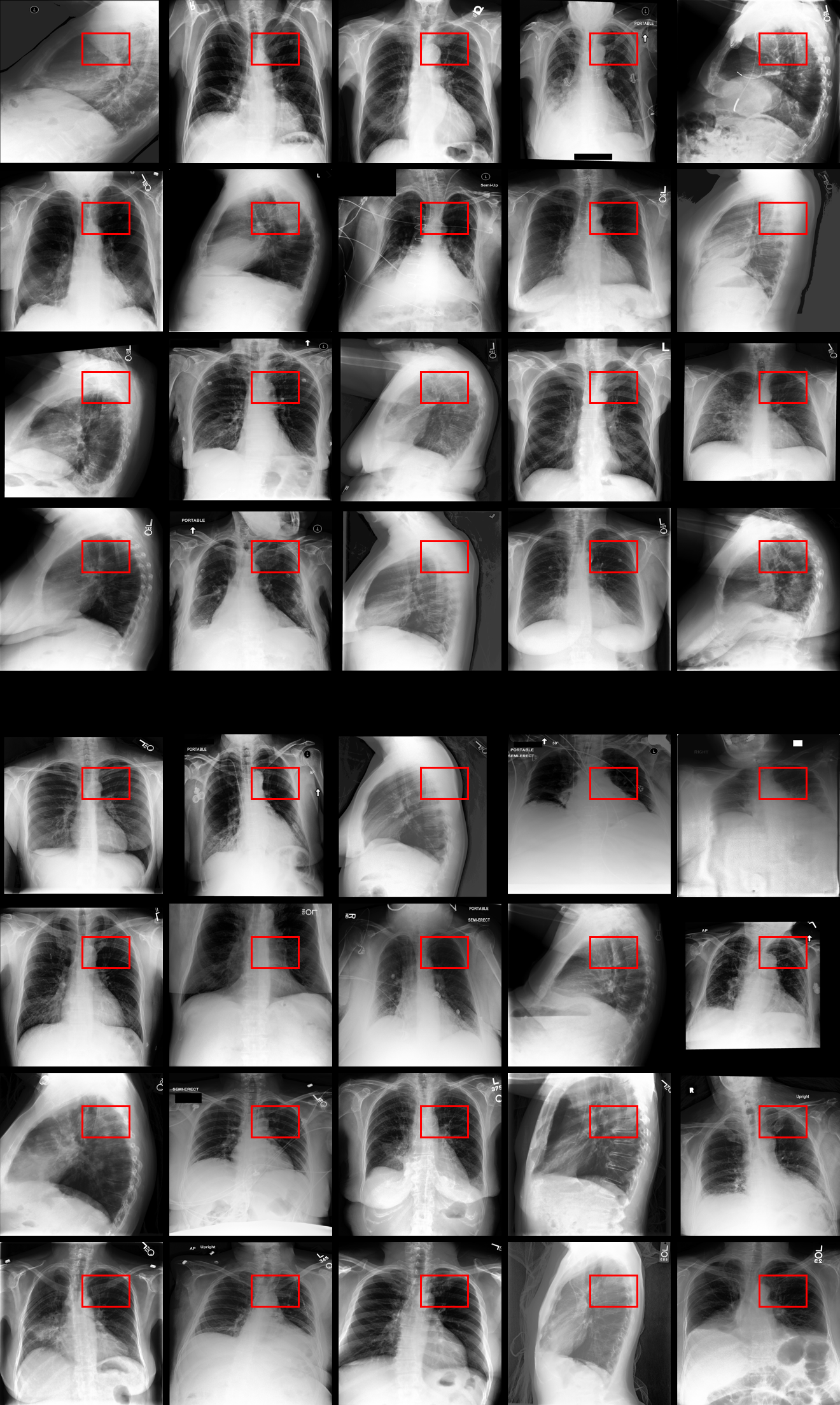}
        \caption*{\seta{}: Pneumonia in right middle lobe.}
    \end{minipage}\hfill
    \begin{minipage}[t]{0.48\linewidth}
        \centering
        % no pneumonia
        \includegraphics[width=\linewidth, trim={0 0 0 1720}, clip]{images/medium_1.png}
        \caption*{\setb{}: No pneumonia.}
    \end{minipage}

    \vspace{6pt}

    % -------- Medium 2 --------
    \begin{minipage}[t]{0.48\linewidth}
        \centering
        % bilateral small pleural effusions
        \includegraphics[width=\linewidth, trim={0 1720 0 0}, clip]{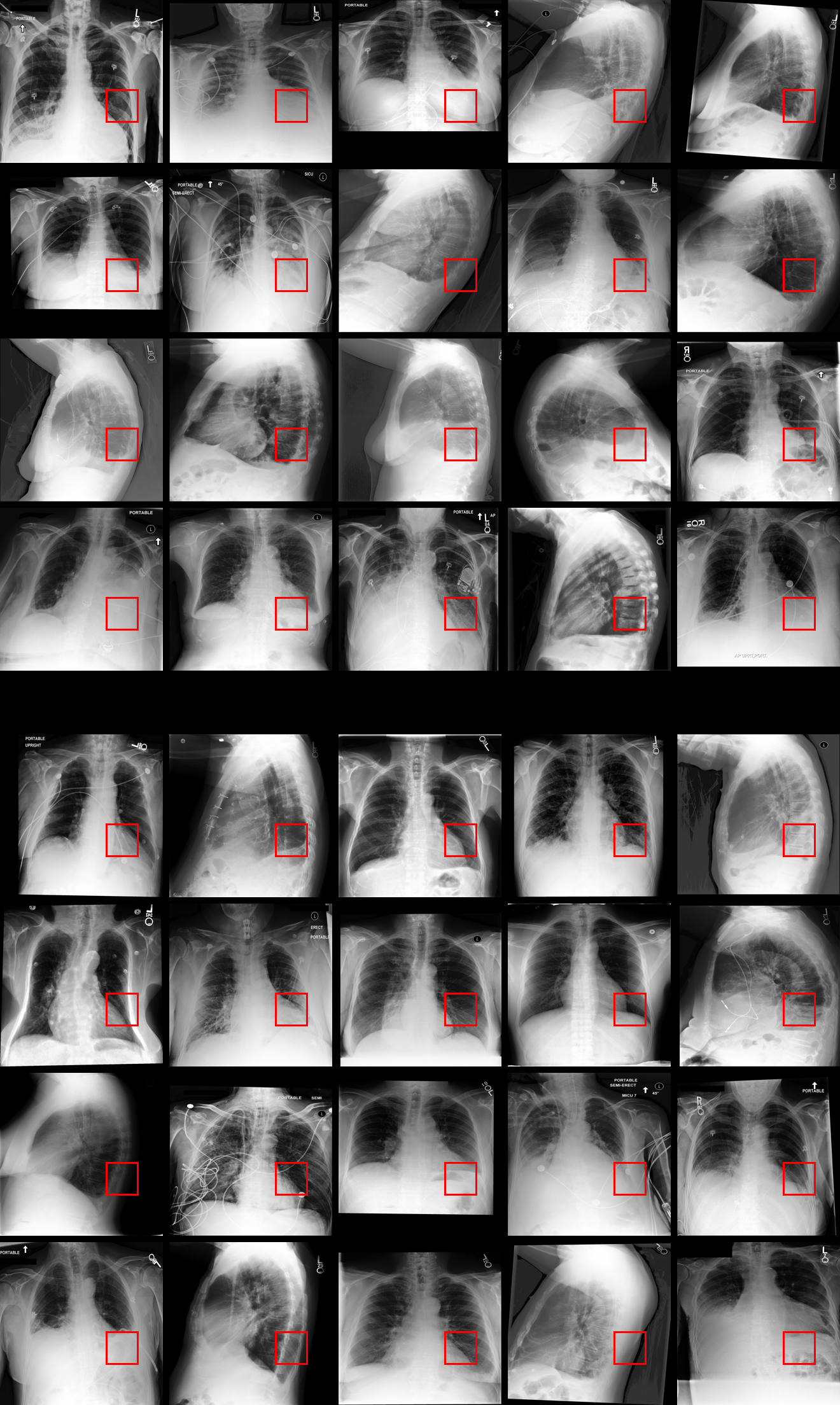}
        \caption*{\seta{}: Bilateral small pleural effusions.}
    \end{minipage}\hfill
    \begin{minipage}[t]{0.48\linewidth}
        \centering
        % no pleural effusion or pneumothorax
        \includegraphics[width=\linewidth, trim={0 0 0 1720}, clip]{images/medium_2.png}
        \caption*{\setb{}: No pleural effusion or pneumothorax.}
    \end{minipage}

    \vspace{6pt}

    % -------- Medium 3 --------
    \begin{minipage}[t]{0.48\linewidth}
        \centering
        % dense right upper lobe airspace consolidation
        \includegraphics[width=\linewidth, trim={0 1720 0 0}, clip]{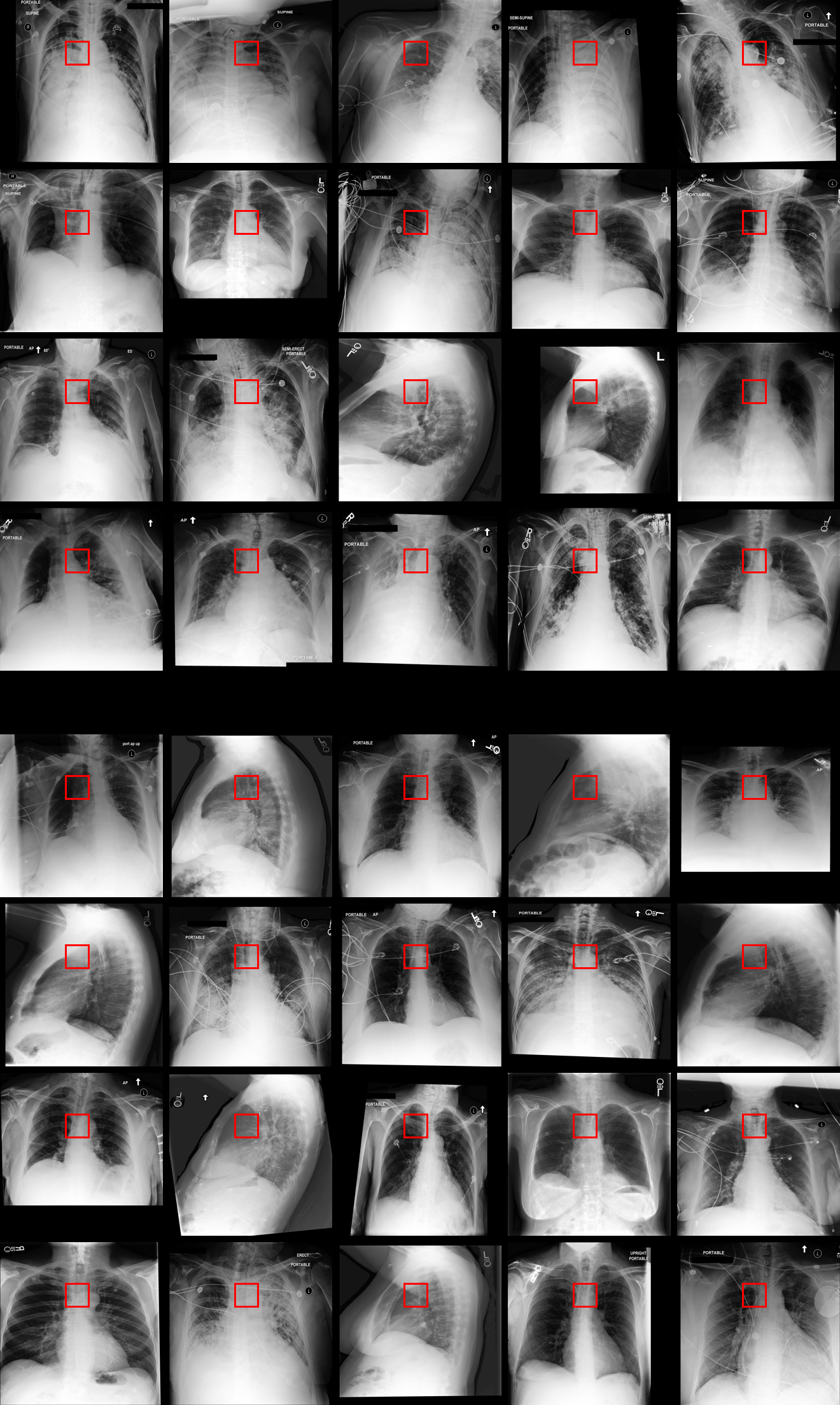}
        \caption*{\seta{}: Dense right upper lobe airspace consolidation.}
    \end{minipage}\hfill
    \begin{minipage}[t]{0.48\linewidth}
        \centering
        % no airspace consolidation
        \includegraphics[width=\linewidth, trim={0 0 0 1720}, clip]{images/medium_3.png}
        \caption*{\setb{}: No airspace consolidation.}
    \end{minipage}

    \caption{\textbf{Medium examples.} We show the set names and the top two differences generated by \method{}. 
    Top row: ``More frequent right lower lobe consolidations suggestive of pneumonia in Group~B" and ``Multifocal pneumonia with opacities in multiple lobes in Group~A." 
    Middle row: ``Presence of pleural effusions with atelectasis and consolidations in Group~A" and ``Bilateral pleural effusions obscuring hemidiaphragms in Group~A."
    Bottom row: ``More extensive bilateral parenchymal opacities in Group~A" and ``Presence of large right pleural effusion with adjacent atelectasis/consolidation in Group~A."}
    \label{fig:medium_examples}
\end{figure*}

\begin{figure*}[!htbp]
    \centering

    % -------- Hard 1 --------
    \begin{minipage}[t]{0.48\linewidth}
        \centering
        % elevated pulmonary venous pressure
        \includegraphics[width=\linewidth, trim={0 1720 0 0}, clip]{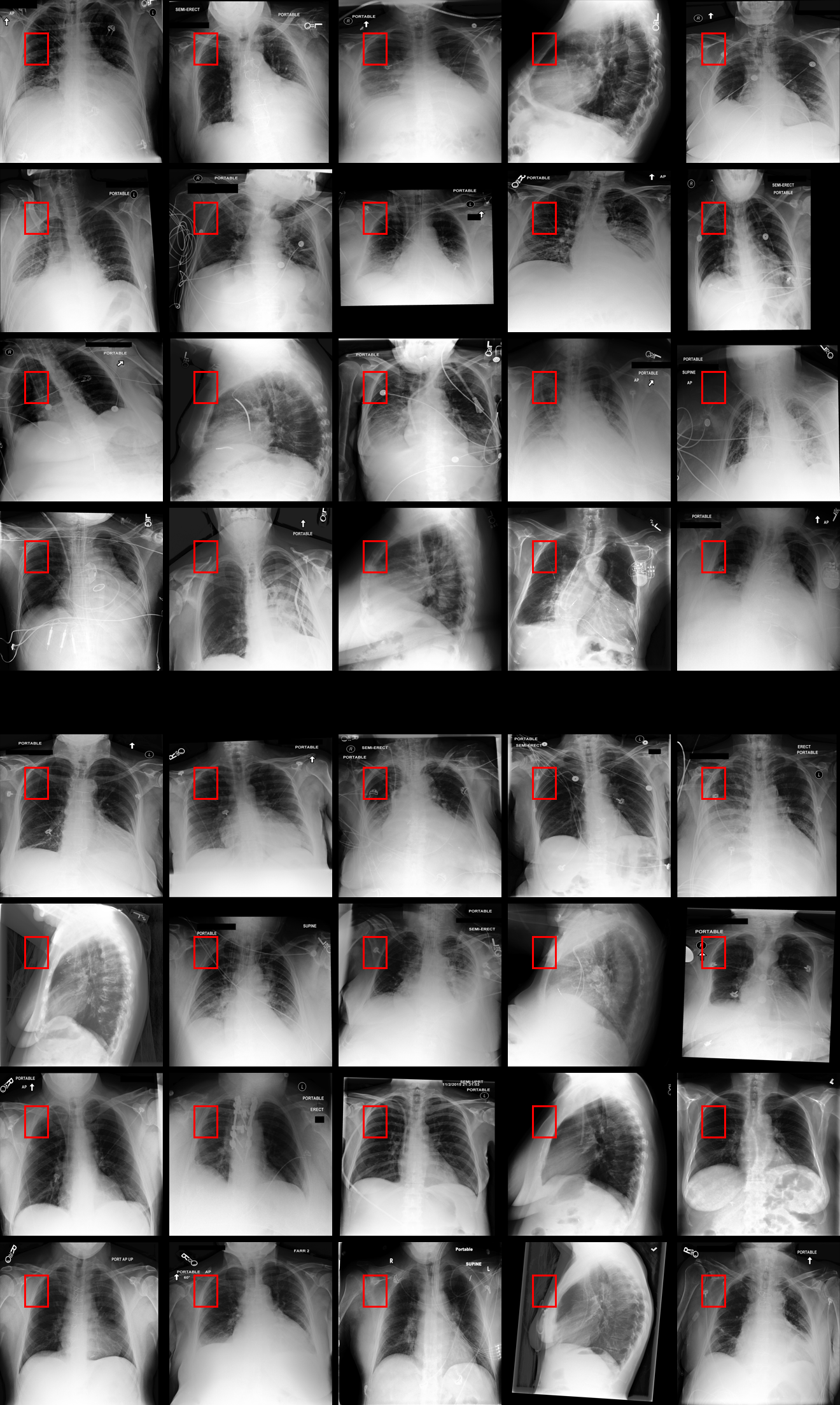}
        \caption*{\seta{}: Elevated pulmonary venous pressure.}
    \end{minipage}\hfill
    \begin{minipage}[t]{0.48\linewidth}
        \centering
        % normal pulmonary venous pressure
        \includegraphics[width=\linewidth, trim={0 0 0 1720}, clip]{images/hard_1.png}
        \caption*{\setb{}: Normal pulmonary venous pressure.}
    \end{minipage}

    \vspace{6pt}

    % -------- Hard 2 --------
    \begin{minipage}[t]{0.48\linewidth}
        \centering
        % normal heart size
        \includegraphics[width=\linewidth, trim={0 1720 0 0}, clip]{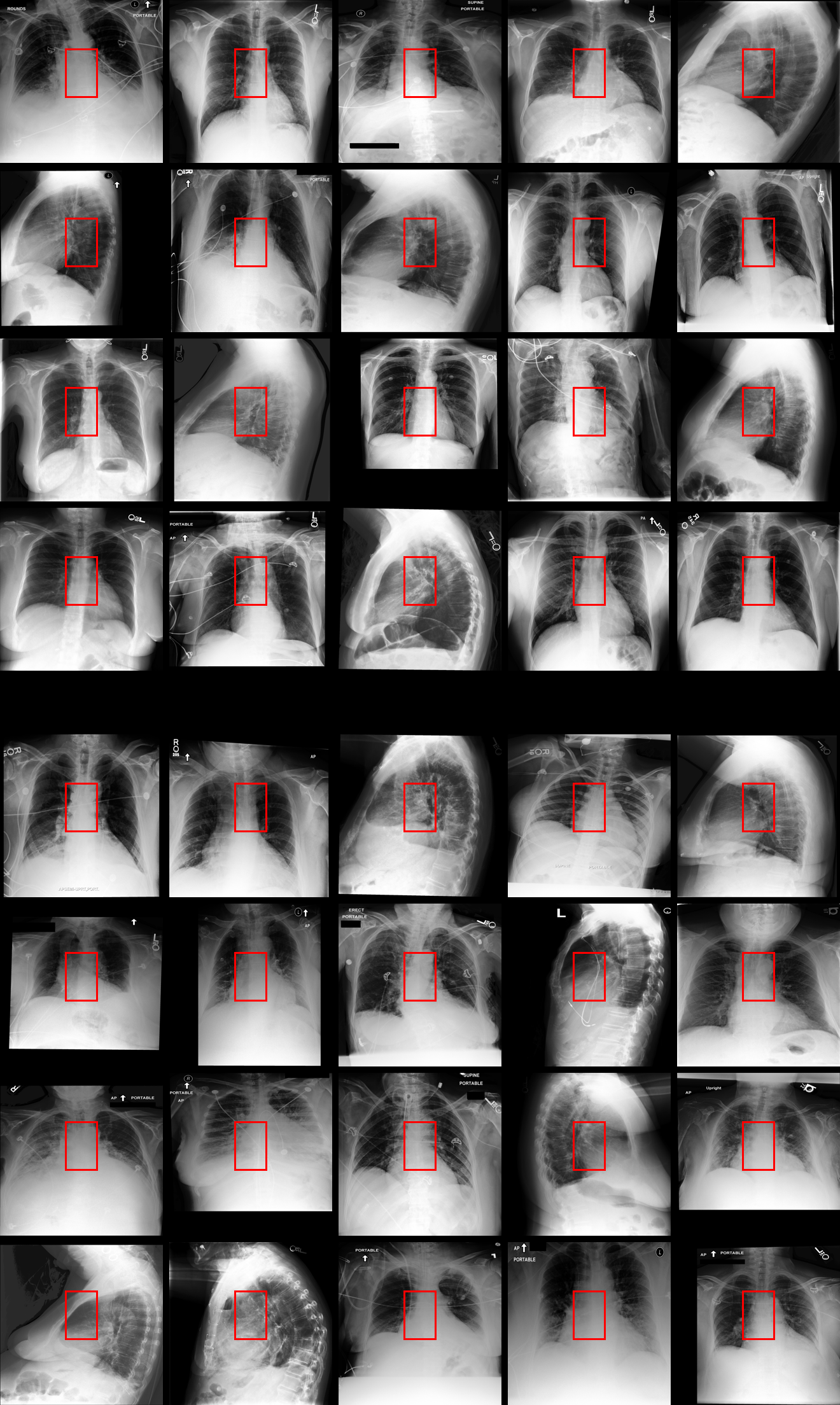}
        \caption*{\seta{}: Normal heart size.}
    \end{minipage}\hfill
    \begin{minipage}[t]{0.48\linewidth}
        \centering
        % enlarged cardiac silhouette
        \includegraphics[width=\linewidth, trim={0 0 0 1720}, clip]{images/hard_2.png}
        \caption*{\setb{}: Enlarged cardiac silhouette.}
    \end{minipage}

    \vspace{6pt}

    % -------- Hard 3 --------
    \begin{minipage}[t]{0.48\linewidth}
        \centering
        % clear basal parenchyma
        \includegraphics[width=\linewidth, trim={0 1720 0 0}, clip]{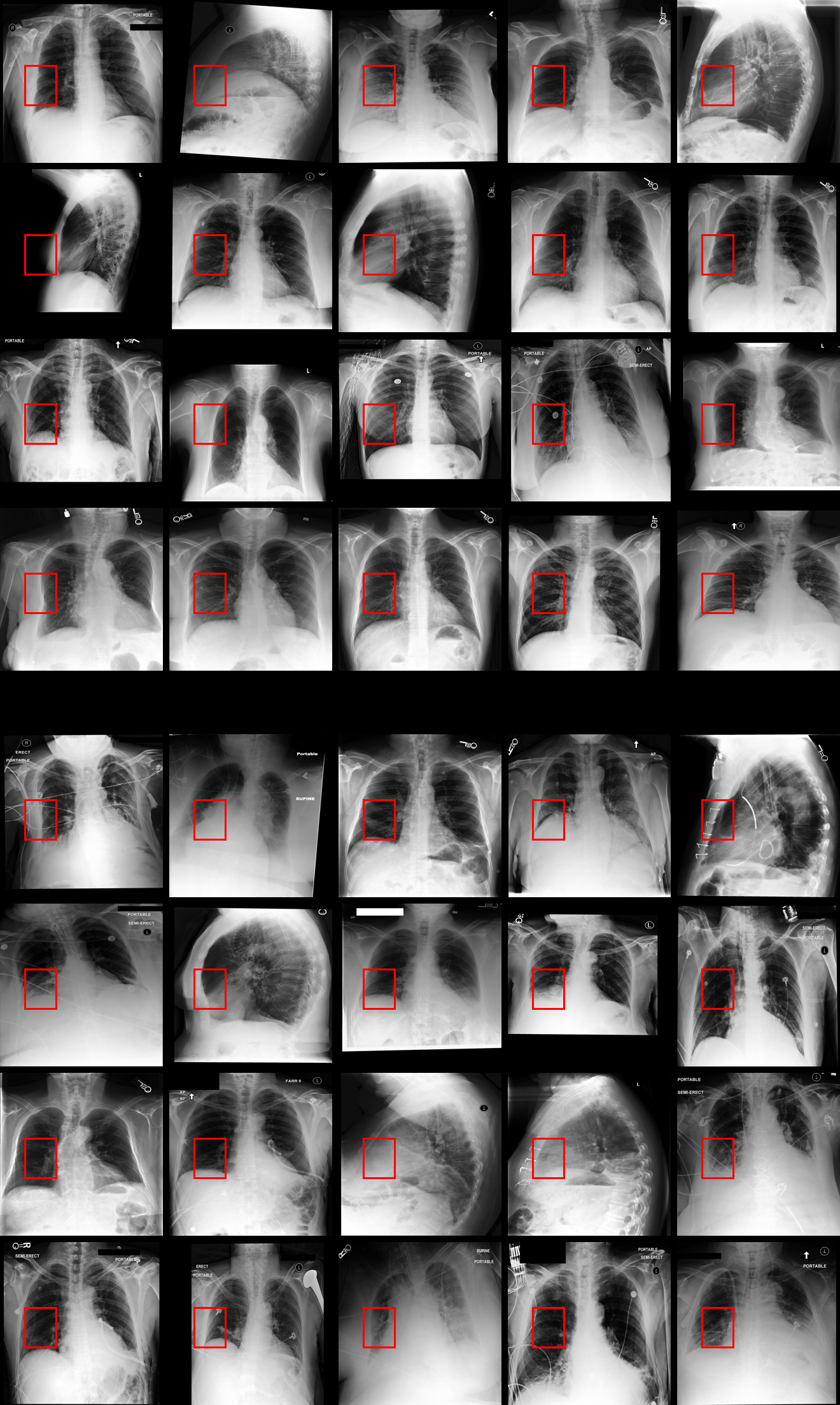}
        \caption*{\seta{}: Clear basal parenchyma.}
    \end{minipage}\hfill
    \begin{minipage}[t]{0.48\linewidth}
        \centering
        % basal atelectasis
        \includegraphics[width=\linewidth, trim={0 0 0 1720}, clip]{images/hard_3.png}
        \caption*{\setb{}: Basal atelectasis.}
    \end{minipage}

    \caption{\textbf{Hard examples.} We show the set names and the top two differences generated by \method{}. 
    Top row: ``More extensive pleural effusions with underlying consolidation in Group~A’’ and ``Pulmonary edema and fluid overload.’’ 
    Middle row: ``More cases with normal heart size and mediastinal/hilar contours in Group~A" and ``Absence of hyperinflation and diaphragm flattening in Group~A." 
    Bottom row: ``Normal cardiomediastinal silhouette and well-expanded lungs in Group~A" and ``Basal atelectasis with associated abnormalities in Group~B."}
    \label{fig:hard_examples}
\end{figure*}

We provide three examples each for Easy/Medium/Hard subset of \benchmark{} in Figures \ref{fig:easy_examples}, \ref{fig:medium_examples}, and \ref{fig:hard_examples}.

\section{Supplementary Section 4}
In this section, we provide additional details of Section 4 in the main paper.

\subsection{Prompts for RadDiff}
We provide the prompts used during multimodal reasoning proposal (Figure \ref{fig:proposal_prompt}), iterative refinement (Figure \ref{fig:iterative_refinement}), and visual search (Figures \ref{fig:coordinates} and \ref{fig:visual_search_prompt}).

\begin{figure*}[!ht]
\centering

% Header bar
\fcolorbox{black}{black}{
  \parbox{0.97\textwidth}{
    \vspace{2pt}
    \color{white}\large\bfseries RadDiff Proposal Prompt 
    \vspace{2pt}
  }
}

\vspace{-1pt}

% Main prompt box
\fcolorbox{black}{gray!5}{
\parbox{0.97\textwidth}{
\small\ttfamily
\begingroup
\obeylines

The following are the results of captioning two groups of chest X-ray images used for a detailed medical analysis:

\{text\}

We also have the two groups of medical chest X-ray images shown below as well. Group A chest X-rays are shown in the first image, while Group B Chest X-rays are part of the second image. 

Your task:

You are the best radiologist in the world. Can you identify the most salient differences between these two groups of chest X-rays, using the above captions and attached images. 

Provide the differences in a clear way (i.e "A has more xxx", but only return "xxx")

Make sure to analyze the captions and images carefully and extract 5-10 salient differences that are more frequently observed in Group A compared to Group B. 

Make sure to only provide information of what group A has more of. 

Don't mention anything about group B in your set of differences. 

Answer with a list of the most distinct salient differences:   

\endgroup
}
}

\caption{Prompt used for multimodal reasoning proposal stage in RadDiff.}
\label{fig:proposal_prompt}
\end{figure*}

\begin{figure*}[!ht]
\centering

% Header bar
\fcolorbox{black}{black}{
  \parbox{0.97\textwidth}{
    \vspace{2pt}
    \color{white}\large\bfseries Iterative Refinement Prompt 
    \vspace{2pt}
  }
}

\vspace{-1pt}

% Main prompt box
\fcolorbox{black}{gray!5}{
\parbox{0.97\textwidth}{
\small\ttfamily
\begingroup
\obeylines

The following are the results of captioning two groups of chest X-ray images used for a detailed medical analysis:

\{text\}

We also have the two groups of medical chest X-ray images shown below as well. Group A chest X-rays are shown in the first image, while Group B Chest X-rays are part of the second image. 

Your task:

You are the best radiologist in the world. Can you identify the most salient differences between these two groups of chest X-rays, using the above captions and attached images. 

Provide the differences in a clear way (i.e "A has more xxx", but only return "xxx")

Make sure to analyze the captions and images carefully and extract 5-10 salient differences that are more frequently observed in Group A compared to Group B. 

Make sure to only provide information of what group A has more of. 

Don't mention anything about group B in your set of differences. 

Here are the top \{top\} differences and scores from the previous round:

\{prev\_results\}

Refine and improve upon these results.

Answer with a list of the most distinct salient differences:

\endgroup
}
}

\caption{Prompt used for Iterative Refinement in RadDiff.}
\label{fig:iterative_refinement}
\end{figure*}

\begin{figure*}[!ht]
\centering

% Header bar
\fcolorbox{black}{black}{
  \parbox{0.97\textwidth}{
    \vspace{2pt}
    \color{white}\large\bfseries Coordinates Query Prompt in Visual Search 
    \vspace{2pt}
  }
}

\vspace{-1pt}

% Main prompt box
\fcolorbox{black}{gray!5}{
\parbox{0.97\textwidth}{
\small\ttfamily
\begingroup
\obeylines

The following are the results of captioning two groups of chest X-ray images used for a detailed medical analysis:

\{text\}

We also have the two groups of medical chest X-ray images shown below as well. Group A chest X-rays are shown in the upper half of the image, while Group B Chest X-rays are part of the lower half of the image.  

Here are the top \{top\} differences and scores from the previous round:

\{prev\_results\}

For each of the top \{top\} findings listed below, we'd like you to pick one area on a chest X-ray image that best shows the difference. 

Please give us a set of four numbers - x1, y1, x2, y2 - that describe a rectangle covering that area. Each number should be between 0 and 1, and they should be based on the size of the image (for example, 0 means the far left or top of the image, and 1 means the far right or bottom). We'll use these rectangles to crop the images and take a closer look at the areas where the differences are most visible and clinically important.

\endgroup
}
}

\caption{Prompt used for Coordinates Query in Visual Search.}
\label{fig:coordinates}
\end{figure*}

\begin{figure*}[!ht]
\centering

% Header bar
\fcolorbox{black}{black}{
  \parbox{0.97\textwidth}{
    \vspace{2pt}
    \color{white}\large\bfseries Visual Search Prompt 
    \vspace{2pt}
  }
}

\vspace{-1pt}

% Main prompt box
\fcolorbox{black}{gray!5}{
\parbox{0.97\textwidth}{
\small\ttfamily
\begingroup
\obeylines

MEDICAL CONTEXT: You are analyzing two distinct cohorts of chest X-ray images for differential diagnostic patterns.

CAPTION ANALYSIS DATA:
\{text\}

VISUAL DATA: The attached images show 5 cropped regions highlighting previously identified differences. Each image has:

- UPPER SECTION (Group A): Separated by a visual gap from Group B

- LOWER SECTION (Group B): Below the visual gap

CLINICAL TASK:

As a board-certified radiologist, perform comparative analysis to identify radiological findings that are statistically more prevalent in Group A.

ANALYSIS REQUIREMENTS:

1. Focus on specific anatomical structures and pathological findings

2. Use precise medical terminology (e.g., "consolidation," "pleural effusion," "cardiomegaly")

3. Consider both caption data and visual evidence

4. Prioritize clinically significant differences

PREVIOUS ITERATION RESULTS:
\{prev\_results\}

REFINEMENT INSTRUCTIONS:

- Enhance specificity of previous findings

- Eliminate false positives or artifacts

- Focus on reproducible patterns across multiple images

- Prioritize diagnostically relevant features

OUTPUT FORMAT:

Provide exactly 5-10 refined findings as single-phrase medical terms (e.g., "bilateral lower lobe consolidation", "enlarged cardiac silhouette", "pleural thickening"):

\endgroup
}
}

\caption{Prompt used for Visual Search in RadDiff.}
\label{fig:visual_search_prompt}
\end{figure*}

\section{Supplementary Section 5}
In this section, we provide additional details of Section 5 in the main paper.

\subsection{Qualitative Analysis}

\textbf{Iterative Refinement.} To illustrate how \method{} converges toward stable, high-confidence differences, we show a representative case comparing \seta{}: heart size be normal and \setb{}: Cardiac silhouette remain enlarged.

\begin{table}[!ht]
\centering
\footnotesize
\setlength{\tabcolsep}{3pt}
\begin{tabular}{c p{0.58\columnwidth} c}
\toprule
\textbf{Rank} & \textbf{Predicted Difference} & \textbf{Score} \\
\midrule
1 & More cases with normal pulmonary vasculature in Group~A & 0.846 \\
2 & More cases showing normal osseous structures & 0.844 \\
3 & More instances of normal heart size and mediastinal/hilar contours & 0.830 \\
4 & More instances of normal heart size and mediastinal contours in Group~A & 0.823 \\
5 & More instances of normal or unchanged cardiomediastinal silhouette & 0.815 \\
\bottomrule
\end{tabular}
\caption{Top candidate differences during first iteration. \method{} finds candidates in different areas; \method{} then uses them to reflect and refine, emphasizing ``normal heart size" in the final iteration.}
\label{tab:iter-refinement}
\end{table}

The model generates several candidates in different areas seen in Table \ref{tab:iter-refinement}. After iterative refinement, the model converges to the ground truth \seta{}: heart size be normal:
\begin{quote}
“\emph{Normal heart size and mediastinal/hilar contours in Group~A},
\emph{More consistent normal findings across Group~A compared to Group~B}"
\end{quote}

\begin{figure*}[!ht]
\centering

\begin{minipage}[t]{0.48\linewidth}
    \centering
    \includegraphics[width=\linewidth]{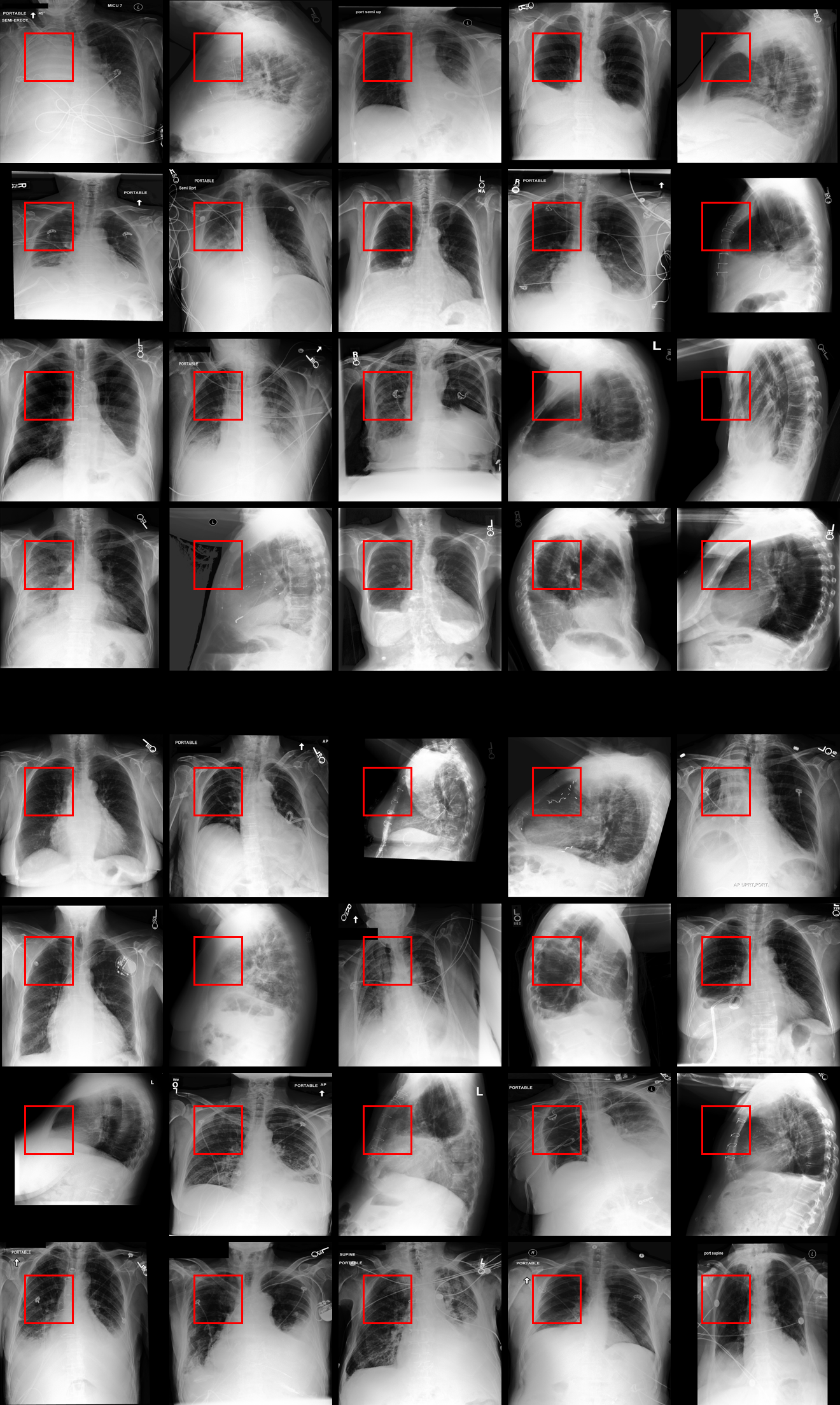}
    \caption*{(a) 
    \method{} progressively zooms into the right lower lung. 
    }
\end{minipage}
\hfill
\begin{minipage}[t]{0.48\linewidth}
    \centering
    \includegraphics[width=\linewidth]{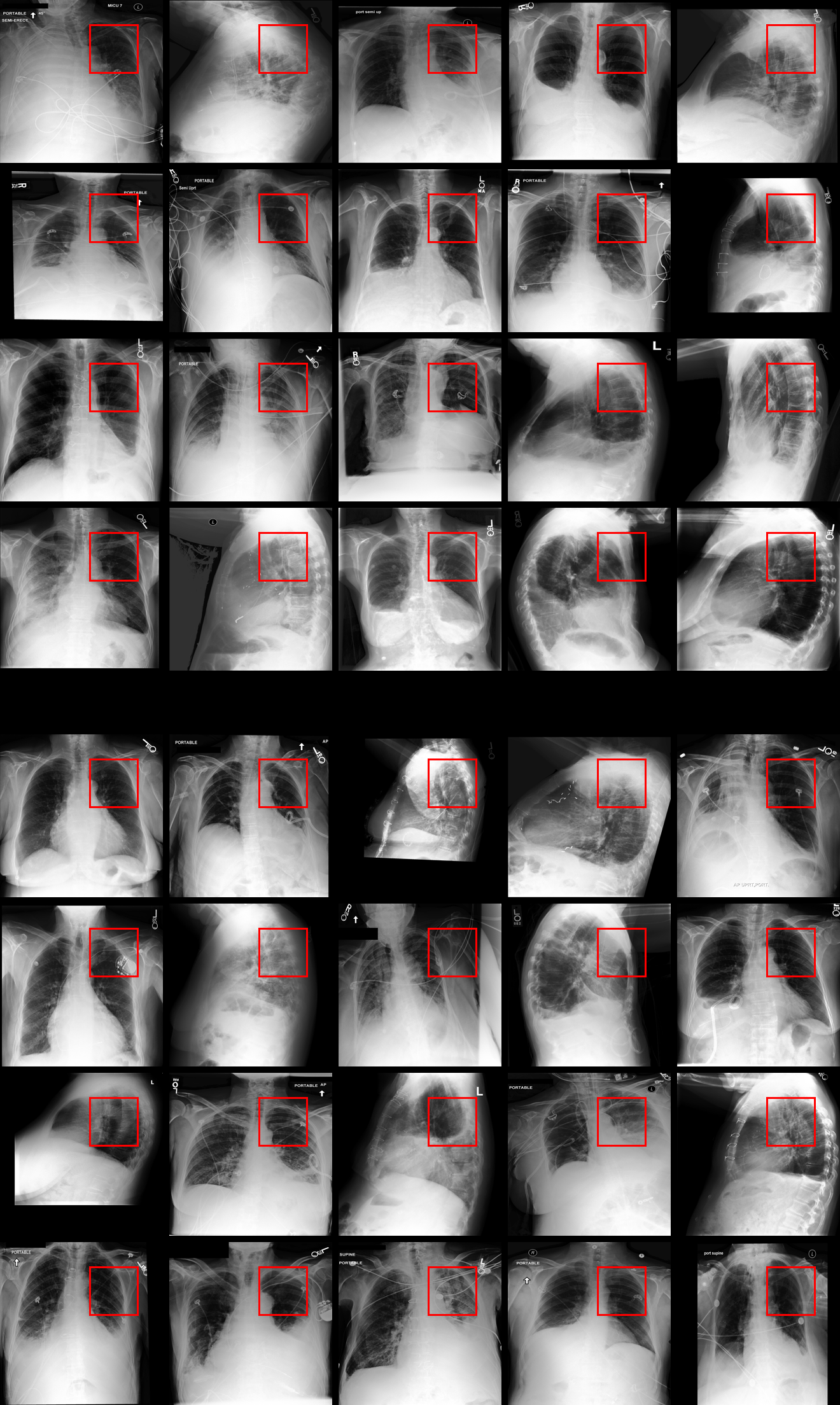}
    \caption*{(b) 
    \method{} scans alternative regions and finds no effusion-related signal.}
\end{minipage}

\caption{\textbf{Visual search.} 
\method{} refines attention toward clinically relevant regions. For example, it progressively zooms into right lower lung to correctly identify more pleural effusion than elsewhere. We display the exact experiment setting here where each figure is an 8$\times$5 grid (40 images): the top half displays \seta{} images and the bottom half displays \setb{} images.}
\label{fig:vs_combined}
\end{figure*}

\textbf{Visual Search.} We then present a qualitative example illustrating how Visual Search refines its focus. When comparing moderate right pleural effusion vs. no pleural effusion, the model’s attention increasingly concentrates on clinically relevant regions, e.g. right lung. This progressive refinement mirrors a radiologist’s iterative inspection process (see Figure~\ref{fig:vs_combined}).

\subsection{RadDiff Case Study}
We now present a detailed qualitative example illustrating how \method{} identifies the most discriminative difference between two sets of radiology images.

\paragraph{Ground-truth distinction.}
\begin{itemize}[leftmargin=1.2em]
    \item \textbf{Group A}: Dense right upper-lobe airspace consolidation.
    \item \textbf{Group B}: No airspace consolidation.
\end{itemize}

\paragraph{First Iteration Top differences proposed by \method{}.}
Table~\ref{tab:cs-proposals} lists the top-ranked differences along with their scores.  

\begin{table}[!ht]
\centering
\footnotesize
\setlength{\tabcolsep}{3pt}
\begin{tabular}{c p{0.58\columnwidth} c}
\toprule
\textbf{Rank} & \textbf{Predicted Difference} & \textbf{Score} \\
\midrule
1 & More extensive bilateral parenchymal opacities & 0.786 \\
2 & More widespread pulmonary opacities indicating multifocal pneumonia / edema & 0.766 \\
3 & More reports of extensive bilateral pulmonary opacities & 0.749 \\
4 & More bilateral pulmonary opacities present & 0.702 \\
5 & Presence of large pleural effusions in Group A & 0.680 \\
\bottomrule
\end{tabular}
\caption{Top 5 proposed differences and alignment scores.}
\label{tab:cs-proposals}
\end{table}

\begin{figure*}[!ht]
  \centering

  % --- Left subplot ---
  \begin{subfigure}[t]{0.48\linewidth}
    \centering
    \includegraphics[width=\linewidth]{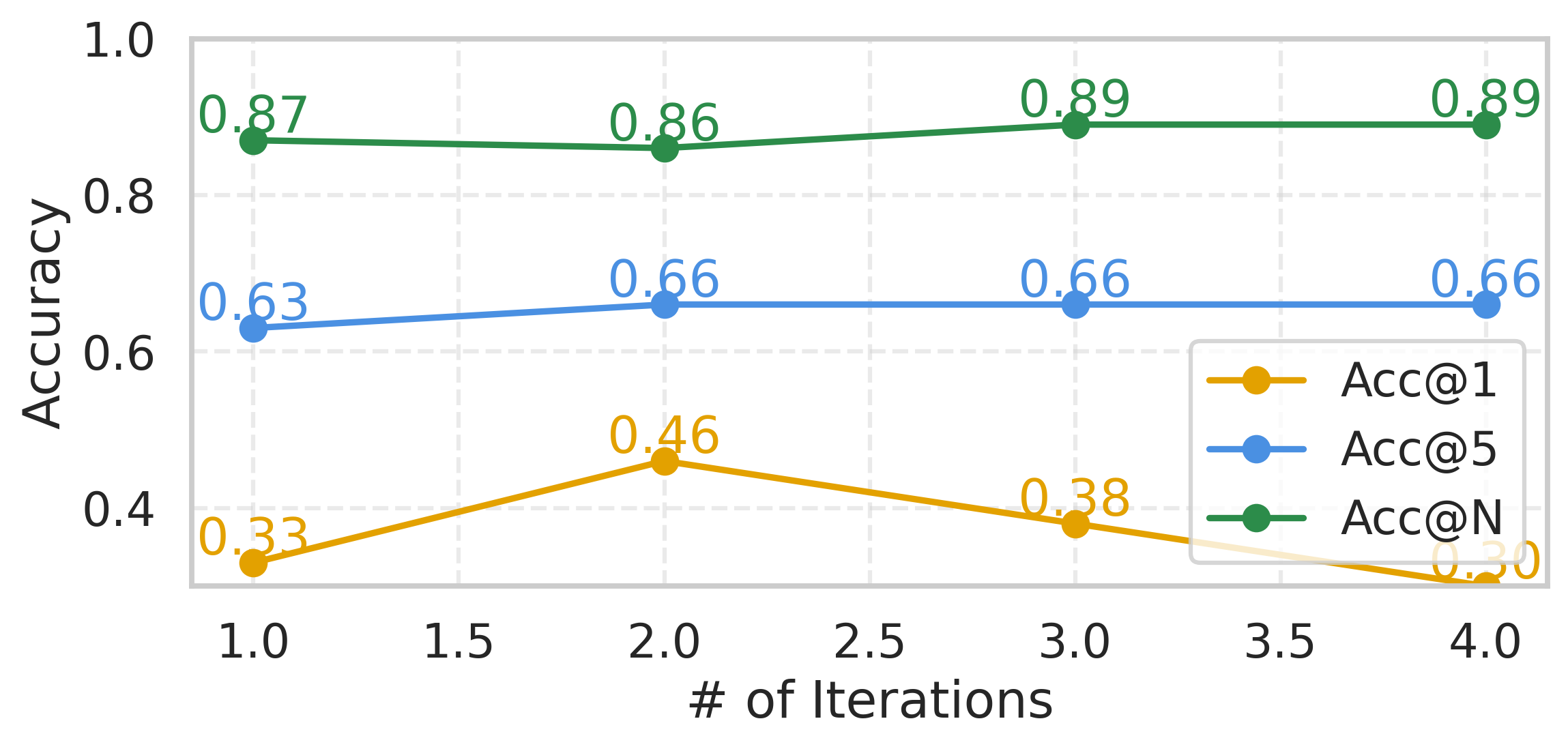}
    \caption*{(a) Model-based iterative refinement.}
  \end{subfigure}
  \hfill
  % --- Right subplot ---
  \begin{subfigure}[t]{0.48\linewidth}
    \centering
    \includegraphics[width=\linewidth]{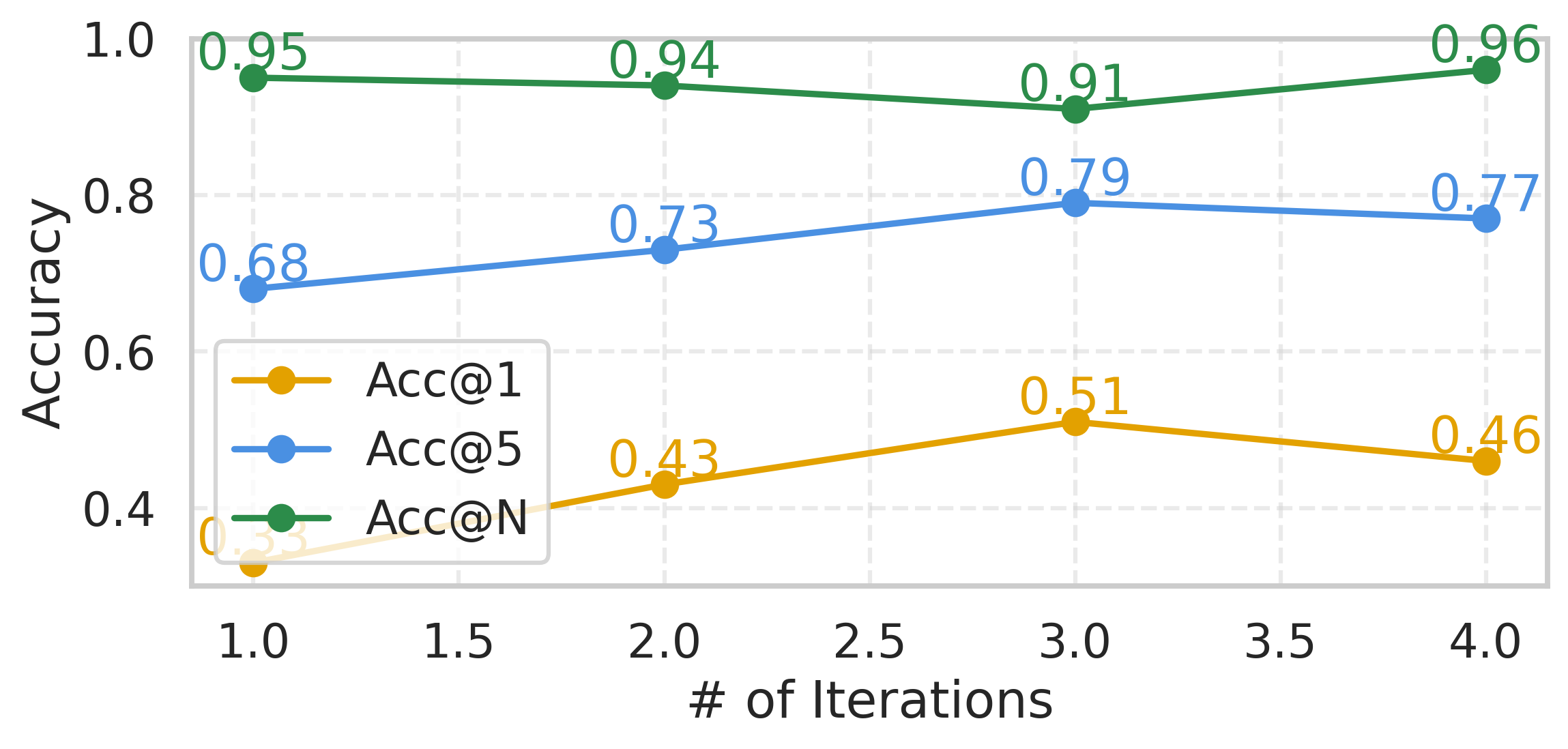}
    \caption*{(b) Ground-truth–based iterative refinement.}
  \end{subfigure}

  \vspace{-0.5em}
  \caption{\textbf{Ablation of iterative refinement rounds.} Both captioner-based and ground-truth–based refinement improve performance, with accuracy plateauing around the second or third refinement round.}
  \label{fig:more_ablation}
\end{figure*}

\paragraph{Refined differences after \method{}'s iterative refinement and visual search.}
After iterations, \method{} produces more anatomically specific and more discriminative statements:

\begin{itemize}[leftmargin=1.2em]
    \item Distribution and extent of lung parenchymal abnormalities favoring large, bilateral consolidations in Group A.
    \item Less extensive bilateral opacities and absence of large pleural effusions in Group~B.
    \item More extensive bilateral parenchymal opacities in Group~A.
    \item More diffuse pulmonary edema pattern with diffuse bilateral opacities in Group~A.
    \item Large pleural effusions with associated atelectasis predominantly in Group~A.
\end{itemize}

\begin{figure*}[!ht]
\centering

% -------- Top 1 --------
\begin{minipage}[t]{0.48\linewidth}
    \centering
    \includegraphics[width=\linewidth, trim={0 1680 0 0}, clip]{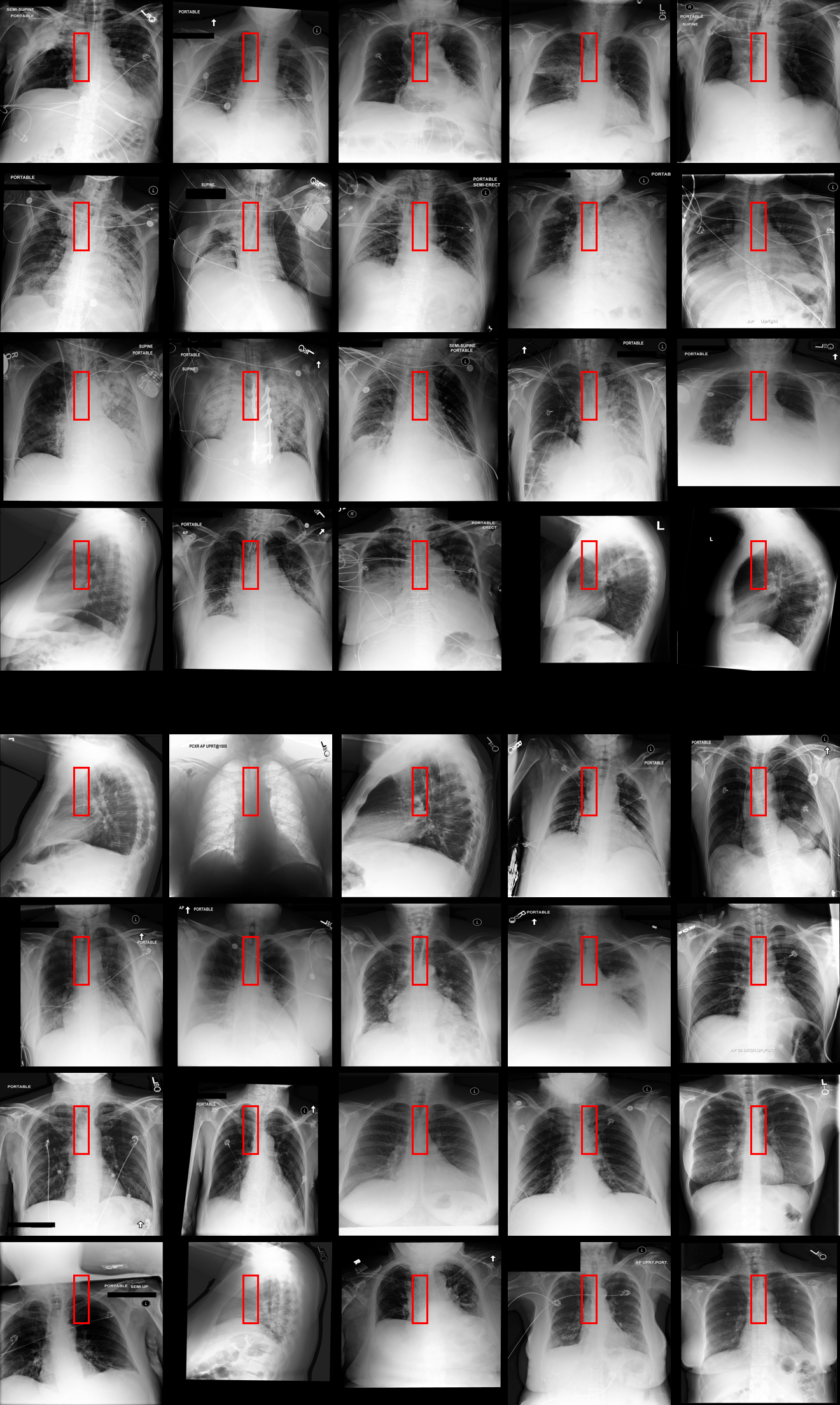}
\end{minipage}\hfill
\begin{minipage}[t]{0.48\linewidth}
    \centering
    \includegraphics[width=\linewidth, trim={0 0 0 1680}, clip]{images/case_study_0.png}
\end{minipage}

\begin{minipage}[t]{0.8\linewidth}
    \centering
    \caption*{(a) Cropped by Top-1 candidate difference: \emph{More extensive bilateral parenchymal opacities}}
\end{minipage}

\vspace{-4pt}

% -------- Top 2 --------
\begin{minipage}[t]{0.48\linewidth}
    \centering
    \includegraphics[width=\linewidth, trim={0 1680 0 0}, clip]{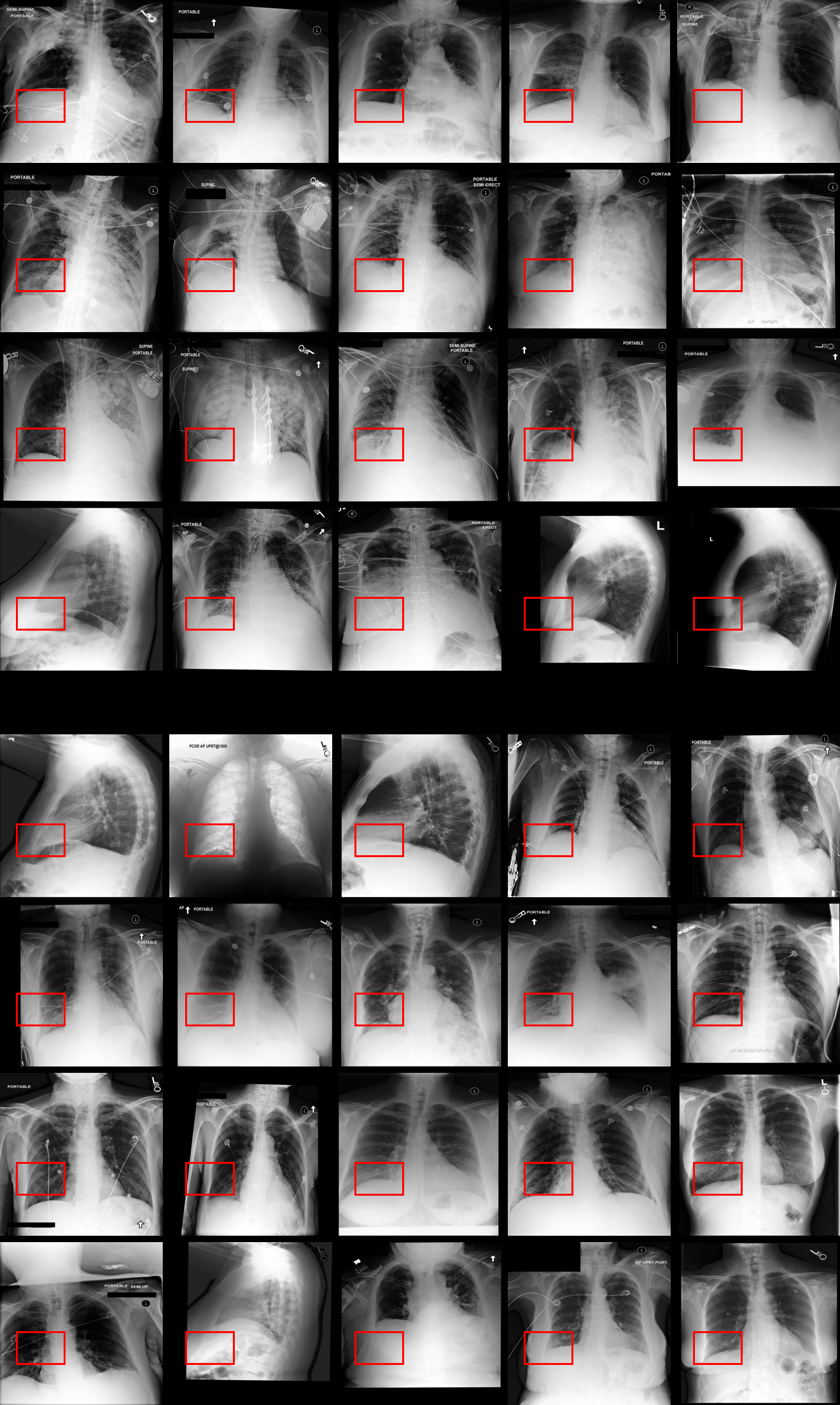}
\end{minipage}\hfill
\begin{minipage}[t]{0.48\linewidth}
    \centering
    \includegraphics[width=\linewidth, trim={0 0 0 1680}, clip]{images/case_study_1.png}
\end{minipage}

\begin{minipage}[t]{1\linewidth}
    \centering
    \caption*{(b) Cropped by Top-2 candidate difference: \emph{More widespread pulmonary opacities indicating multifocal pneumonia / edema}}
\end{minipage}

\vspace{-4pt}

% -------- Top 3 --------
\begin{minipage}[t]{0.48\linewidth}
    \centering
    \includegraphics[width=\linewidth, trim={0 1680 0 0}, clip]{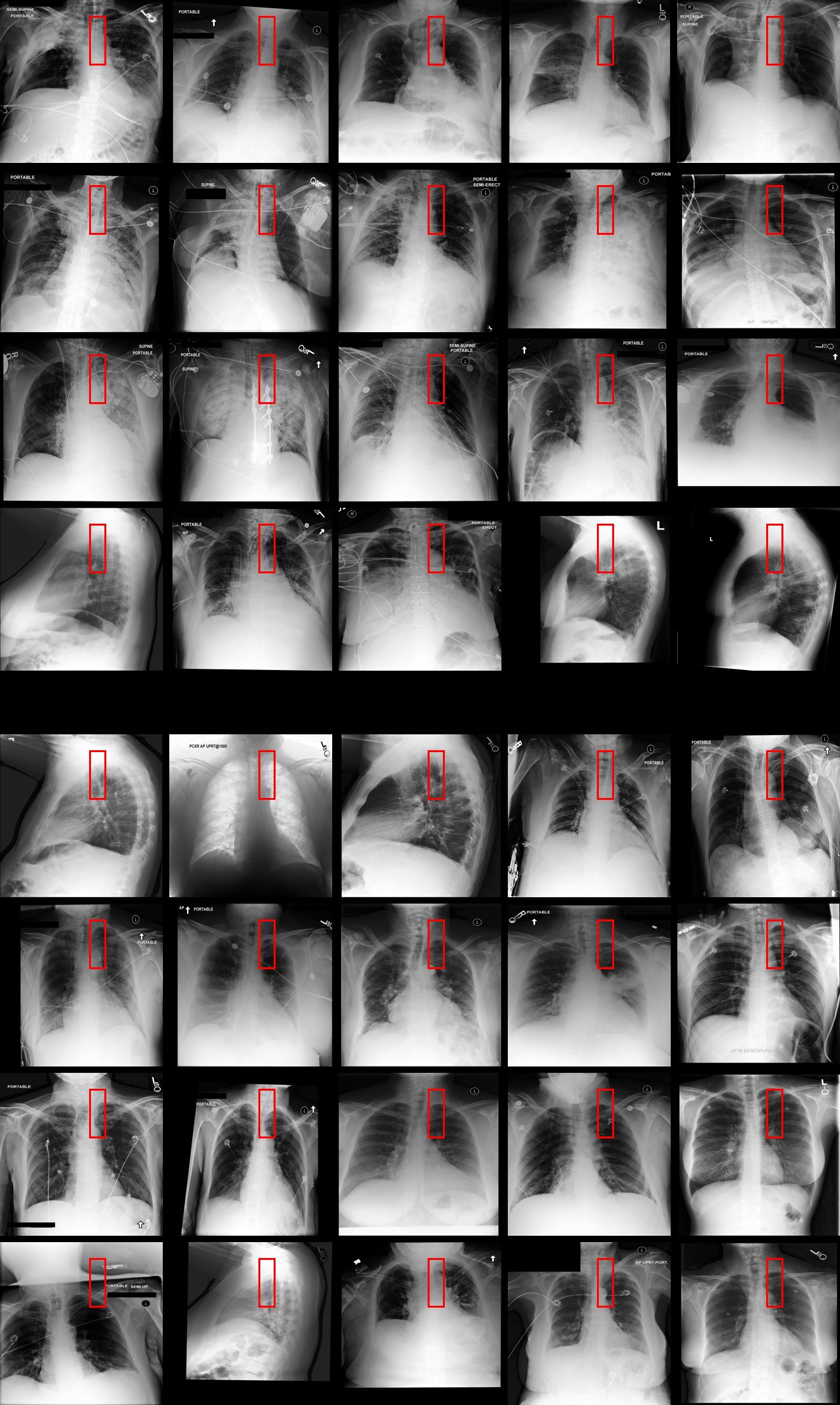}
\end{minipage}\hfill
\begin{minipage}[t]{0.48\linewidth}
    \centering
    \includegraphics[width=\linewidth, trim={0 0 0 1680}, clip]{images/case_study_2.png}
\end{minipage}

\begin{minipage}[t]{0.8\linewidth}
    \centering
    \caption*{(c) Cropped by Top-3 candidate difference: \emph{More reports of extensive bilateral pulmonary opacities}}
\end{minipage}

\vspace{-4pt}

% -------- Top 4 --------
\begin{minipage}[t]{0.48\linewidth}
    \centering
    \includegraphics[width=\linewidth, trim={0 1680 0 0}, clip]{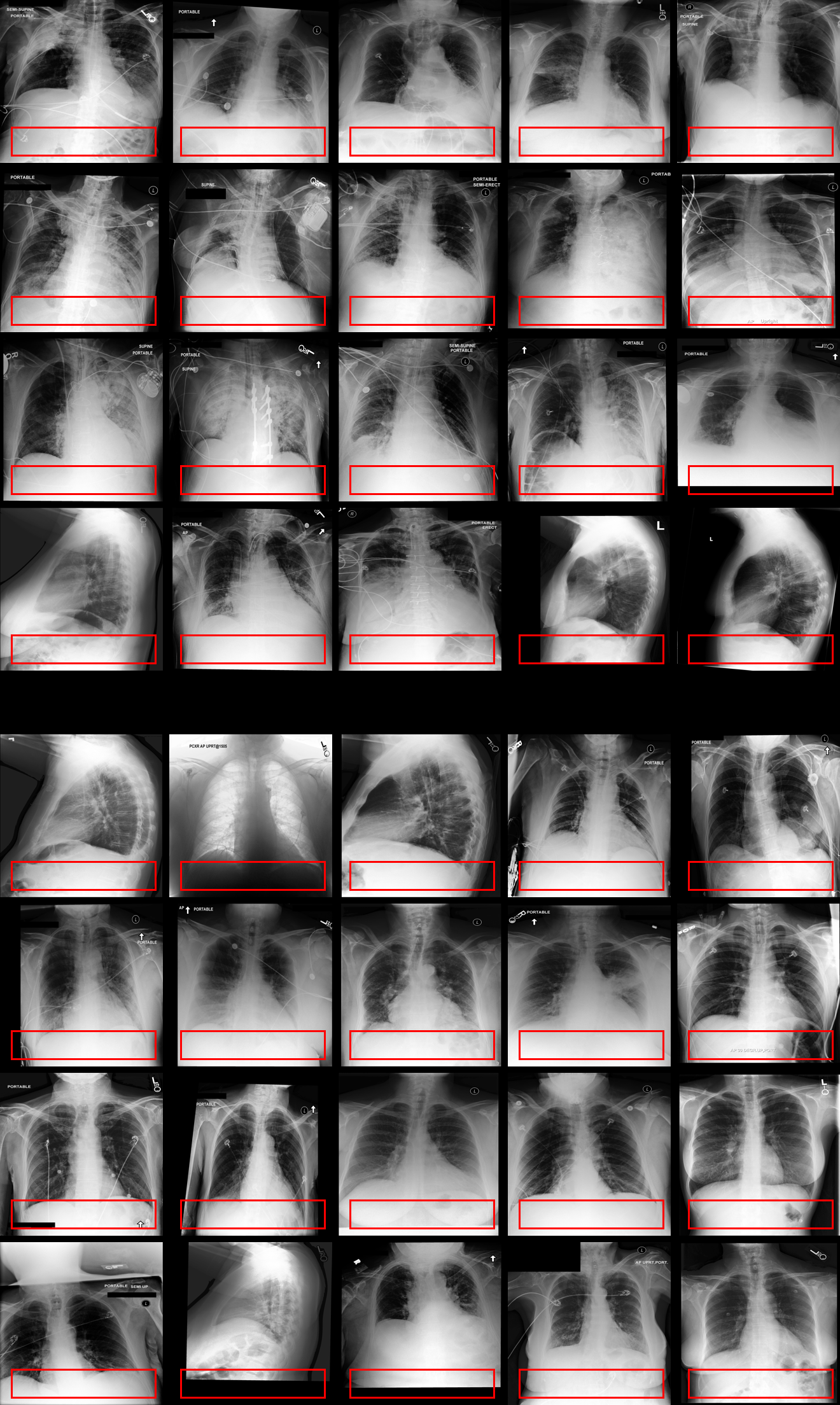}
\end{minipage}\hfill
\begin{minipage}[t]{0.48\linewidth}
    \centering
    \includegraphics[width=\linewidth, trim={0 0 0 1680}, clip]{images/case_study_3.png}
\end{minipage}

\begin{minipage}[t]{0.8\linewidth}
    \centering
    \caption*{(d) Cropped by Top-4 candidate difference: \emph{More bilateral pulmonary opacities present}}
\end{minipage}

\vspace{-4pt}

% -------- Top 5 --------
\begin{minipage}[t]{0.48\linewidth}
    \centering
    \includegraphics[width=\linewidth, trim={0 1680 0 0}, clip]{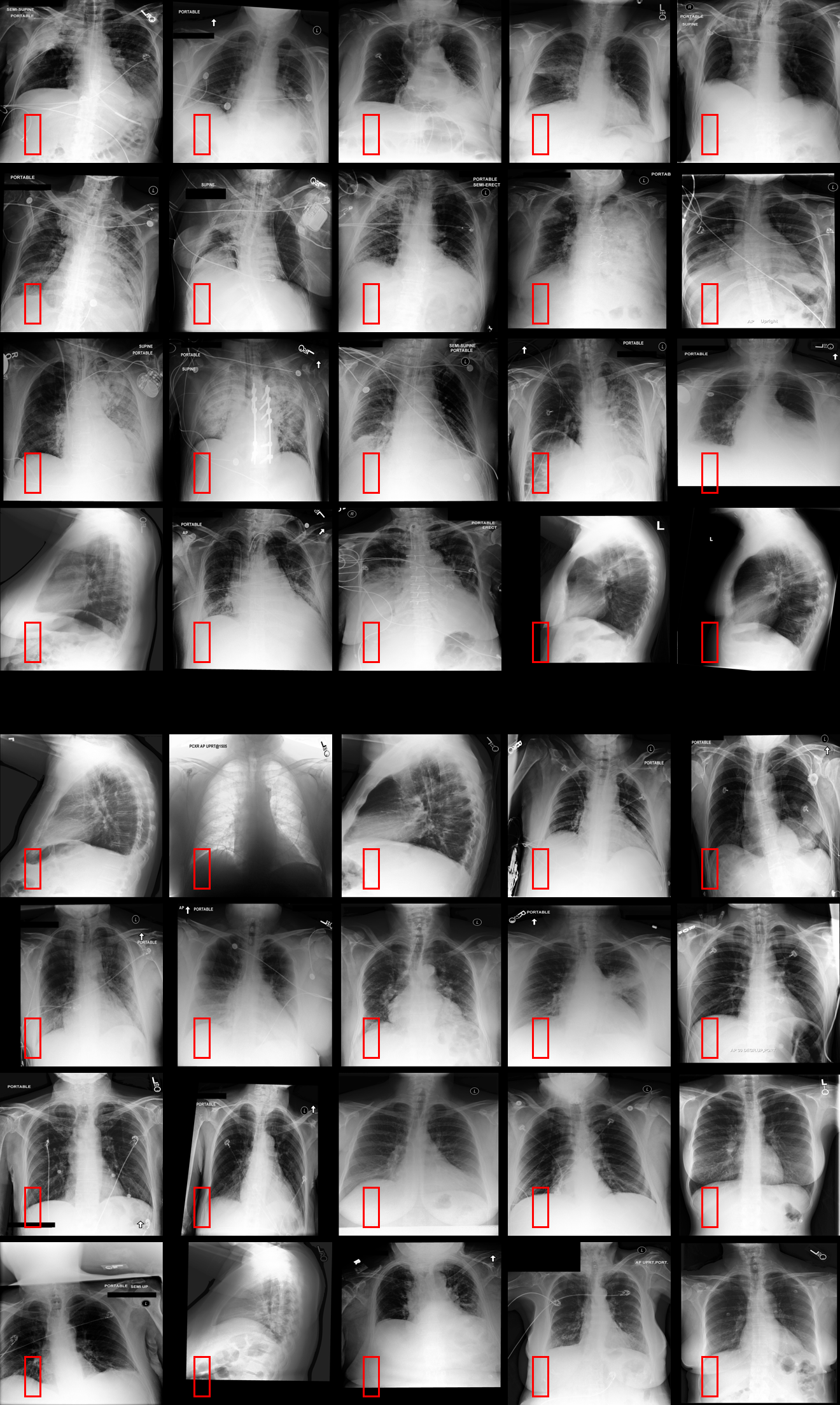}
\end{minipage}\hfill
\begin{minipage}[t]{0.48\linewidth}
    \centering
    \includegraphics[width=\linewidth, trim={0 0 0 1680}, clip]{images/case_study_4.png}
\end{minipage}

\begin{minipage}[t]{0.8\linewidth}
    \centering
    \caption*{(e) Cropped by Top-5 candidate difference: \emph{Presence of large pleural effusions in Group~A}}
\end{minipage}
\vspace{-8pt}
\caption{Visual Search Cropped Areas.
For each Top-K difference, \method{} highlights one corresponded region. The left side shows \seta{}, and right side shows \setb{}.}
\label{fig:vs_top5}
\end{figure*}

\paragraph{Visual Search visualizations.}
Figure \ref{fig:vs_top5} demonstrates how Visual Search works by focusing on regions corresponding to the top five proposed differences from the last iteration.

\subsection{More Ablation Analysis}
Figure \ref{fig:more_ablation} examines how iterative refinement behaves under two conditions: (a) using model-generated captions as input (top-10 candidates), and (b) using ground-truth reports summary during input.

For the top-10 refinement, performance improves from iteration 1 to iteration 2 across Acc@1/5/N, but then plateaus or declines slightly with further iterations. This pattern suggests that early iterations introduce genuinely new refinements, while deeper iterations provide diminishing returns as the candidate pool becomes saturated with recycled or overlapping differences. Consistent with this trend, we report top-10 performance at iteration 2 and top-5 at iteration 3 in the main results.

For the ground-truth–based refinement, accuracy continues to increase through iteration 3 before stabilizing, similar to model-generated captions.

\section{Supplementary Section 6}
In this section, we provide additional details of Section 6 in the main paper.

\subsection{Disease Categories for Application Experiments}
We provide the disease categories table (\ref{tab:disease_codes}) which we use to filter pneumonia cases and COVID-19 cases for the application experiments. 

\begin{table}[ht]
\centering
\scriptsize
\begin{tabular}{lll}
\toprule
\textbf{Disease} & \textbf{ICD Codes} & \textbf{Description} \\
\midrule
Pneumonia & 480--486; J13--J18, J851 & All types \\
Heart Failure & 428*; I50* & Congestive HF \\
COPD & 490--496; J40--J44 & Chronic lung disease \\
Resp. Failure & 518*; J96* & Acute/chronic \\
Sepsis & 99591–99592; A41*, R652* & Septic states \\
ARDS & 51882; J80 & Acute distress \\
\bottomrule
\end{tabular}
\caption{Disease categories and ICD codes used for patient selection.}
\label{tab:disease_codes}
\end{table}

\subsection{Additional Application Results}

We present extended application results for two settings: (1) survival analysis, specifically pneumonia patients who died within 90 days versus who died in hospital, and (2) racial differences, in particular, Asian versus White and White versus Black. \method{} generates a ranked list of candidate differences for each comparison. Below, we provide the full set of candidate differences produced by \method{} for completeness.

\textbf{Pneumonia (90-day death vs.\ died in hospital).}
In the main text, our survival analysis showed results for in-hospital mortality against long-term survivors, revealing clear differences in device burden and intervention-related findings (e.g., tubes, lines, catheters).
We share additional results for the 90-day mortality vs.\ in-hospital mortality comparison. Rather than medical device burden, the candidate differences surface parenchymal patterns (hyperinflation, atelectasis, effusion severity), providing a complementary view of survival-related radiology imaging differences.
We list the full set of candidate differences below for completeness.

\begin{itemize}[leftmargin=*]
\item Hyperinflation with diaphragmatic flattening significantly more common.
\item Hyperinflated lungs with more prominent diaphragmatic flattening.
\item Presence of hyperinflation and flattening of the hemidiaphragms.
\item Smaller or absent pleural effusions.
\item Less evidence of pulmonary fibrosis features.
\item Normal cardiomediastinal silhouette without cardiomegaly.
\item Enlarged mediastinal silhouette noted in some cases.
\item Emphysema with low lung volumes and flattened diaphragms versus extensive bilateral opacities and pneumonia.
\item Bibasilar atelectasis versus more localized or extensive atelectasis.
\item Frequent bibasilar and bilateral atelectasis.
\item Less frequent pulmonary edema or focal pneumonia.
\item More normal pulmonary vasculature and mediastinal contours versus mild edema or cardiomegaly.
\item Bilateral pulmonary hyperinflation more frequent.
\item Focal consolidation and localized pneumothorax more frequent in some cases.
\item Normal cardiomediastinal silhouette versus cardiomegaly with retrocardiac atelectasis and mild edema.
\item Moderate left pleural effusion with adjacent atelectasis.
\item Small pneumothorax more frequently noted.
\item Small right apical pneumothorax versus small-to-moderate contralateral pneumothorax.
\item Compressed or shifted mediastinum.
\item Bilateral pleural effusions with bibasilar atelectasis more prevalent.
\item Presence of right-sided central line with tip at cavoatrial junction.
\item Interval removal of chest tubes without pneumothorax.
\item More instances of small pneumothorax.
\item Small right pleural effusion with overlying atelectasis versus moderate effusions with compressive atelectasis.
\item Absence of bilateral parenchymal opacities and pneumonia versus extensive bilateral opacities.
\item Widespread bilateral parenchymal opacities more frequent.
\item Extensive bilateral parenchymal opacities indicating edema or atelectasis.
\item Absence of widespread pulmonary opacities and consolidations in some cases.
\end{itemize}

\textbf{Asian versus White race comparison.}
In the main text, we compared White vs.\ Asian cohorts and observed a clear underdiagnosis pattern: White patients showed more abnormalities, while Asian patients were more often labeled as normal.
Below, we provide the full list of candidate differences produced by \method{} for Asian versus White. 

\begin{itemize}[leftmargin=*]
\item Small right apical pneumothorax more frequently observed.
\item More cases showing no focal consolidation or pneumothorax.
\item More images with well-expanded lungs.
\item More mentions of a large right upper-lobe mass with air bronchograms.
\item Repeated absence of focal consolidation or pneumothorax.
\item More normal overall findings with occasional complications.
\item More normal mediastinal and hilar contours.
\item Increased lung volumes, including low lung volumes.
\item Normal lung volumes with some low-volume cases.
\item More normal lung findings despite complications.
\item More normal lung findings overall with limited complications.
\item Less frequent emphysema with flattened hemidiaphragms.
\item More normal heart size and vascular structures.
\item Normal heart size and vasculature more common.
\item More frequent large right pleural effusion with atelectasis.
\item More low lung volumes or hyperinflation.
\item More cases without pleural effusion or pneumothorax, except isolated ones.
\item More complications such as pneumonia or loculated effusions.
\item More low-volume lungs with bibasilar atelectasis.
\item More mentions of interventions such as endotracheal tubes or catheters.
\item Fewer cases with large hiatal hernia or major abnormalities.
\item More small left pleural effusions with atelectasis.
\item Higher frequency of small bilateral pleural effusions.
\item Normal cardiomediastinal silhouette with vascular congestion and edema more frequently described.
\end{itemize}

\textbf{White vs.\ Black cohorts.}
We provide the results for comparing White vs.\ Black cohorts. The observed differences follow the same qualitative pattern as in the White–Asian comparison.

The full set of candidate differences is listed below.

\begin{itemize}[leftmargin=*]
\item Higher prevalence of pleural effusions and pneumothorax.
\item Presence of endotracheal tube tips at or above the clavicles noted more often.
\item More cases of small bilateral pleural effusions with atelectasis.
\item More cases with pulmonary vascular congestion and interstitial edema.
\item More frequent projection of tubes over the stomach or above the carina.
\item Bilateral small pleural effusions with overlying atelectasis.
\item Pleural effusions, including large unilateral or left-sided effusions.
\item Tips of endotracheal and nasogastric tubes projecting over the stomach or above the carina.
\item Large or unilateral left-sided pleural effusions more prevalent.
\item Presence of a large hiatal hernia more frequently described.
\item Reports of large hiatal hernia appearing in multiple cases.
\item Small residual pneumothorax (left or right) noted more commonly.
\item Hyperinflated lungs with flattening of the hemidiaphragms (emphysema).
\item Presence of a hiatal hernia.
\item Enlarged cardiac silhouette and signs of pulmonary edema more frequently observed.
\item Greater incidence of hyperinflation with flattening of the hemidiaphragms.
\item Consolidation in the left upper lobe more characteristic in some cases.
\item Enlarged cardiomediastinal silhouette more commonly described.
\item Lower lung volumes with accentuated cardiomediastinal silhouette.
\end{itemize}

\end{document}